\documentclass{article}

\usepackage{PRIMEarxiv}

\usepackage[utf8]{inputenc} 
\usepackage[T1]{fontenc}    
\usepackage{hyperref}       
\usepackage{url}            
\usepackage{booktabs}       
\usepackage{amsfonts}       
\usepackage{nicefrac}       
\usepackage{microtype}      
\usepackage{lipsum}
\usepackage{fancyhdr}       
\usepackage{graphicx}       
\graphicspath{{media/}}     
\usepackage{caption}
\usepackage[export]{adjustbox}
\usepackage[table,xcdraw,dvipsnames]{xcolor}
\usepackage{float}
\usepackage{setspace} 
\usepackage{lineno}
\usepackage{rotating}
\usepackage{graphicx}
\usepackage{subfigure}
\usepackage{multirow}
\usepackage{amsmath}
\usepackage{makecell}
\usepackage[flushleft]{threeparttable}

\usepackage{soul}
\usepackage[toc,page]{appendix}

\usepackage{lmodern,textcomp}

\title{Machine Learning in management of precautionary closures caused by lipophilic biotoxins
\thanks{\textit{\underline{Citation}}: 
\textbf{Molares-Ulloa, A., Fernandez-Blanco, E., Pazos, A., \& Rivero, D. (2022). Machine learning in management of precautionary closures caused by lipophilic biotoxins. Computers and Electronics in Agriculture, 197, 106956. DOI:10.1016/j.compag.2022.106956.}}
}
\author{
  Andres Molares-Ulloa, Enrique Fernandez-Blanco, Alejandro Pazos, Daniel Rivero \\
  Universidade da Coruña, Department of Computer Science and Information Technology, \\
  Faculty of Computer Science, 15071, A Coruña, Spain \\
  Centro de investigación CITIC, Department of Computer Science and Information Technology, \\ 
  University of A Coruña, 15071, A Coruña, Spain \\
  Biomedical Research Institute of A Coruña (INIBIC), University Hospital Complex of A Coruna (CHUAC), \\
  A Coruna, 15006, Spain\\
  \texttt{andres.molares@udc.es} \\
}

\begin{document}
\maketitle

\begin{abstract}
Mussel farming is one of the most important aquaculture industries. The main risk to mussel farming is harmful algal blooms (HABs), which pose a risk to human consumption. In Galicia, the Spanish main producer of cultivated mussels, the opening and closing of the production areas is controlled by a monitoring program. In addition to the closures resulting from the presence of toxicity exceeding the legal threshold, in the absence of a confirmatory sampling and the existence of risk factors, precautionary closures may be applied. These decisions are made by experts without the support or formalisation of the experience on which they are based. Therefore, this work proposes a predictive model capable of supporting the application of precautionary closures. Achieving sensitivity, accuracy and kappa index values of 97.34\%, 91.83\% and 0.75 respectively, the kNN algorithm has provided the best results. This allows the creation of a system capable of helping in complex situations where forecast errors are more common.
\end{abstract}

\keywords{Machine Learning \and Harmful Algal Blooms \and Mussels \and Aquaculture \and  Diarrhoeic Shellfish Poisoning}

\section{Introduction}
\label{S:1}

Global mussel production has steadily increased to 2.2 million tonnes in 2018, more than double the amount produced ten years ago (\cite{fao}). Nearly 94\% of global mussel production comes from aquaculture (\cite{avdelas2021decline}). Young mussels are harvested from the sea and may be grown on suspended ropes; these ropes, which are covered with mussel seed held in place with nylon nets, are suspended either from rafts, or wooden frames, or from longlines with floating plastic buoys. A substantial portion of EU production is farmed on suspended ropes, a technique that can be extended further offshore and which, although very sensitive to plankton blooms, is the only one that could allow further increases in production.

One of the main risks of mussel farming is Harmful Algal Blooms (HABs). HABs are episodes of high concentrations of algae, including some cyanobacteria and microalgae that are potentially toxic for human consumption. This is because there is a risk of poisoning by consuming filter-feeding bivalve molluscs such as mussels that feed on these algae, accumulating the toxins in their meat. To monitor these episodes, there are programs set up in mussel production areas. For the early detection of high toxicity events, these monitoring programmes have fixed sampling points strategically located in the production areas. These high toxicity events can lead to a temporary suspension of mussel harvesting and marketing. The most common toxin-producing species are those of Diarrhoeic Shellfish Poisoning (DSP) type. The most abundant of which is the dinoflagellate \textit{Dinophysis acuminata}) (\cite{vilas2008ria}).

The opening and closing of the production areas is based on the analysis of the toxicity of the mollusc meat, as established by European legislation (\cite{UE627/2019}). Within the monitoring programme, sampling planning uses expert knowledge based on information on endogenous and exogenous factors influencing the proliferation of potentially toxic phytoplankton species. The most compromising point of this process is the absence of sampling during non-working days or when inclement weather does not allow it to be carried out. This leads to situations where it is impossible to collect the data to support an effective closure. If there are indications of a potential increase in toxicity levels, the competent authority is legally entitled to proceed with 'precautionary closures' of bivalve mollusc production areas.

Precautionary closures may become effective after a subsequent analysis verifying the presence of toxins, otherwise, the closure will be lifted. The application or non-application of these measures creates two possible problem scenarios. In the first scenario the precautionary closure is applied even though toxicity values above the legal threshold are not reached. This scenario could lead to economic losses for producers because they are prohibited from working while the area remains closed. In the second scenario no indications of a high toxicity event are detected, but a subsequent analysis shows the presence of toxins. The latter is a much more dangerous situation than the previous one because, during this period of extraction activity, there is a potential risk of introducing contaminated shellfish into the market, with the consequent risk to public health. Today, the implementation of precautionary closures is based on the experience of monitoring experts. The existence of a predictive model could help them make the right decisions in complex situations.

Harmful algal blooms are not only a potential risk to public health, they are also a major problem for the production sector. Work such as that of Di Jin and Porter Hoagland (\cite{jin2008value}) has shown that the development of predictive systems can lead to significant improvements in management strategy and profits for the farming sector. So far, numerous studies have attempted such predictions around the world, notably off the coasts of South Korea (\cite{lee2018improved}), Hong Kong (\cite{yu2021predicting}) and the Persian Gulf (\cite{ebrahimi2015algae}), in general, these works have focused their efforts on predicting biomarkers such as the concentration of toxic phytoplankton in the water or chlorophyll-a (\cite{deng2021machine,liu2009modeling}). There are studies for the specific case of the Spanish coast (\cite{VeloSurez2007ArtificialNN}) and specifically for the Galician coast (\cite{vilas2014support,aguilar2017prediccion,molares2020}). For the creation of this type of predictive models, the use of different classical techniques has been compared with ML techniques to try to find the co-figuration that best suits this problem (\cite{cruz2021review,liu2009modeling}). It was determined that ML techniques outperform classical methods. The success of applying machine learning techniques to harmful algal blooms lies in the selection of the relevant data and the pre-processing of the data. The proliferation of harmful algal blooms is influenced by many factors, the most important of which are: temperature, water flow, upwelling, light, nutrients and salinity.

A higher water temperature favours algae proliferation, as well as thermocline stratification favours their concentration (\cite{davis2009effects}). Excessive water flow and circulation disperses algae concentrations, reducing the occurrence of blooms (\cite{li2013effect}). The light is necessary for phytoplankton to photosynthesise (\cite{paerl2012climate}). Dissolved nutrients in the water create a favourable environment for algal growth (\cite{paerl2012climate}). The salinity plays an important role in the formation of phytoplankton communities (\cite{gasiunaite2005seasonality}).

The best results were obtained using the combined CNN and LSTM spatio-temporal classification technique to classify and discriminate between HAB and non-HAB events produced in Florida coastal waters by the algae \textit{Karenia brevis} (\cite{hill2020habnet}). But it is difficult to have such a large volume of data on a regular basis, and even impossible for many regions. Therefore, we have studied the effect of sample size (\cite{guallar2016artificial}) and modelling with feature reduction (\cite{rahman2013algae}). 

As mentioned above, chlorophyll-a concentration is one of the most recurrent biomarkers of potentially toxic phytoplankton proliferation (\cite{rahman2013algae}). Chlorophyll-a is related to the concentration of phytoplankton containing this pigment, but not only biotoxin-producing phytoplankton contain it. Therefore, this biomarker may be in error when algal blooms are of non-harmful algae. On the other hand, if the objective is to close mussel production areas as a result of exceeding the legal threshold for the presence of biotoxins (\cite{molares2020}), this could lead to a significant improvement in the accuracy of the prediction.

\begin{figure}[h]
  \centering
  \includegraphics[width=0.9\textwidth]{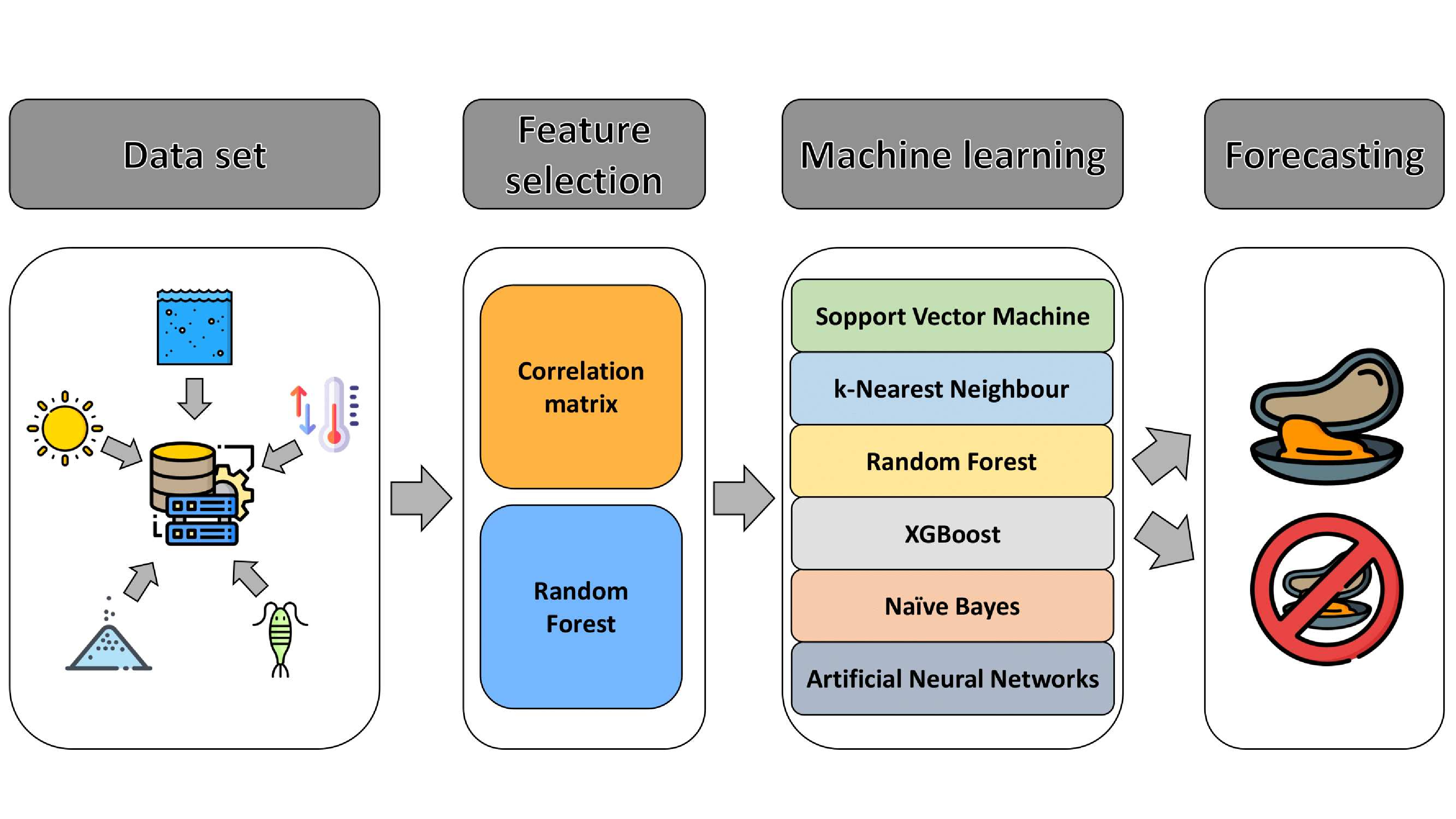}
  \caption{Schematic representation of the machine learning-based system for predicting harmful algal bloom closures and aiding decision making in mussel farming. This graphic has been designed with resources from Flaticon.com}
  \label{fig:workflow}
\end{figure}

For this reason, the objective of this study is the creation of a predictive model of high toxicity events in mussel production areas. Consequently, the classification of mussel production areas will focus on whether the presence of lipophilic toxin in mussel flesh exceeds the legal threshold or not. To do this, a comparison of solutions will be carried out, applying different machine learning techniques to predict the state of production areas affected by DSP-type toxins. Taking into account previous studies carried out in the field (\cite{cruz2021review}), a total of 6 classification techniques were selected: Artificial Neural Network (ANN), Support Vector Machines (SVMs), k-Nearest Neighbour (kNN), XGBoost, Random Forest and Na\"ive Bayes. This model can be used by government agencies with responsibilities in the control of shellfish production areas and its use can be of benefit to the mussel industry and the consumer. A workflow of the proposed system can be seen in figure \ref{fig:workflow}.

The structure of this paper is defined as follows: It starts with a section on advances in the field of HAB prediction, and in particular in the use of ML techniques for this purpose. In section \ref{sec:mm} the techniques used as well as the configuration of the techniques used are presented. The results of these models can be found in section \ref{sec:r} and will be interpreted in section \ref{sec:d}. Finally, in sections \ref{sec:c} and \ref{sec:fw} the conclusions obtained and the lines of future work are presented.

\section{Materials and Methods} 
\label{sec:mm}
\subsection{Dataset and its construction}

\begin{figure}[h]
  \centering
  \includegraphics[width=1\textwidth]{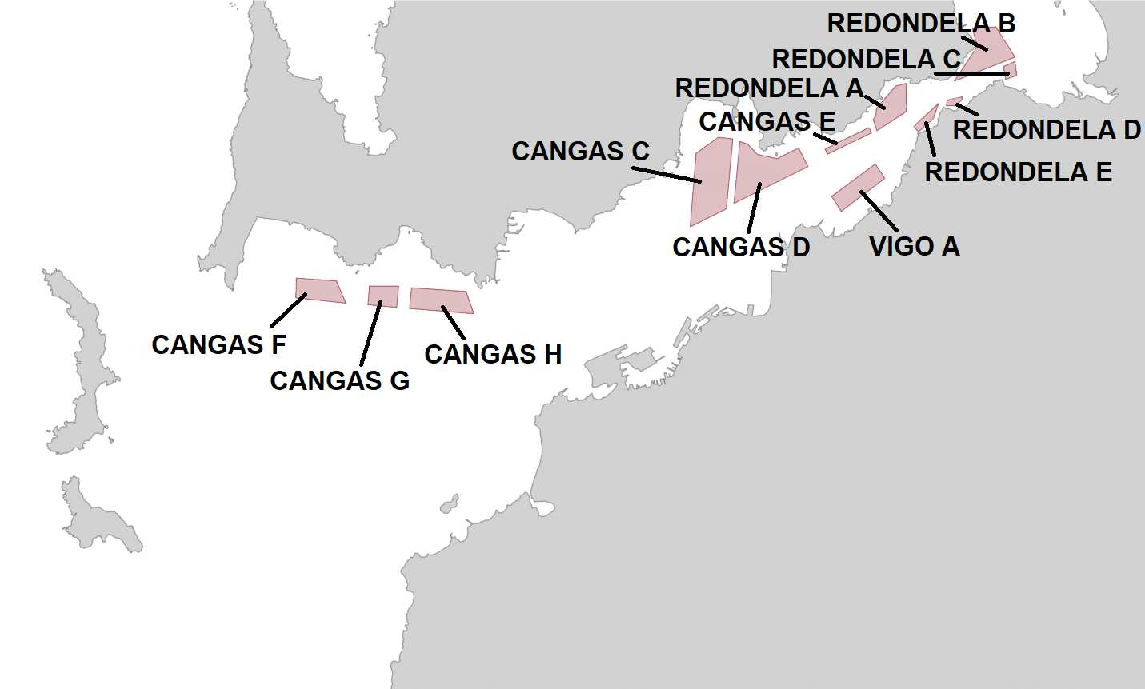}
  \caption{Map of the production areas of cultivated molluscs in the Vigo estuary. Source: \url{http://193.144.46.136/EstadoZonas/Default.aspx?tmapa=0}}
  \label{fig:zonas}
\end{figure}

\begin{figure}[h]
  \centering
  \includegraphics[width=1\textwidth]{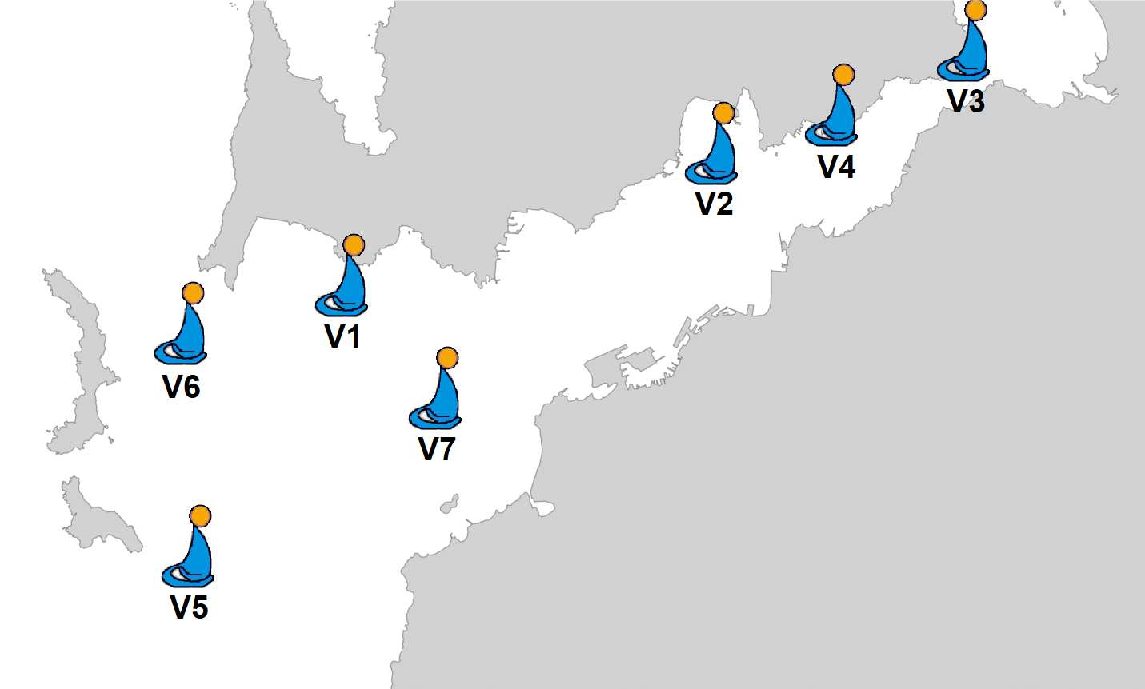}
  \caption{Map of oceanographic stations located in the Vigo estuary. Source: \url{http://www.intecmar.gal/Ctd/Default.aspx}}
  \label{fig:estaciones}
\end{figure}

The production status (open/closed) of the crop areas has been used as the target variable. This status of crop areas is assigned according to whether or not the presence of toxin in the mussel tissue exceeds the legal threshold. If the threshold is exceeded, extraction activities in that crop area will cease (closure of the crop area) or, if not, extraction activity will be allowed (opening of the crop area). It was decided to focus the study on predicting the state of the cultivation areas each Monday. This is because no toxin presence analysis is carried out on the previous days (Saturday and Sunday), which is one of the most compromised points of the existing monitoring system. Twelve out of thirteen mussel production areas of the Vigo estuary (Galician coast, Spain) have been selected: Cangas F, Cangas G, Cangas H, Cangas C, Cangas D, Cangas E, Vigo A, Redondela A, Redondela B, Redondela C, Redondela D and Redondela E (see figure \ref{fig:zonas}), excluding from this study the production area of Baiona A because it is a polygon that remains unsampled for long periods of time. As the areas are managed independently, and as input variables, we have used a set of environmental and oceanographic data of different nature, recorded by different institutions between 2004 and 2018. The network of sampling points for phytoplankton monitoring coincides, to a large extent, with the stations set up to determine oceanographic conditions, figure \ref{fig:zonas}. Weekly, an oceanographic vessel takes samples from points V1, V2, V3, V4, V5, V6 and V7, located in the Vigo estuary. Their distribution can be seen in the figure \ref{fig:estaciones}). In each sampling point, integrated samples of water between 0 and 15 metres deep to count phytoplankton cells and determine nutrients dissolved in water, were taken. Simultaneously, a multiparametric probe measures the physico-chemical parameters of the water column. The different variables collected in these oceanographic stations, as well as other constant variables for the whole estuary obtained thanks to METEOGALICIA (\cite{meteogalicia}) and the IEO (\cite{raia}), are shown in the table \ref{tab:variables}. All oceanographic stations have been taken into account in order to know which ones offer the data most related to the occurrence and concentration of HAB, as this depends directly on the functioning of factors such as the morphological configuration of the estuary itself or sea currents. By analysing the data collected, it is determined that the sampling frequency of the data collected is mainly weekly, so this metric will be used as a reference for the creation of the models.

\begin{table}[h]
\centering
\resizebox{1\textwidth}{!}{%
\begin{tabular}{|c|c|c|c|c|}
\hline
\multirow{2}{*}{\textbf{Source}} & \multirow{2}{*}{\textbf{Variable}} & \textbf{Number of} & {\textbf{Features}} & \multirow{2}{*}{\textbf{Frequency}} \\
                                       &                                                & \textbf{locations}    & \textbf{generated}  &         \\ \hline
\multirow{10}{*}{\textbf{INTECMAR}}    & Temperature                                    & 7                     & 14                 & Weekly \\
                                       & Salinity                                       & 7                     & 7                  & Weekly \\
                                       & Oxygen                                         & 7                     & 7                  & Weekly \\
                                       & Chlorophyll-a concentration                & 7                     & 7                  & Weekly \\
                                       & \textit{Dinophysis acuminata} cells abundance  & 7                     & 7                  & Weekly \\
                                       & Dissolved ammonium                             & 7                     & 7                  & Weekly \\
                                       & Dissolved phosphate                            & 7                     & 7                  & Weekly \\
                                       & Dissolved nitrate                              & 7                     & 7                  & Weekly \\
                                       & Dissolved nitrite                              & 7                     & 7                  & Weekly \\
                                       & State of production areas                      & 1                     & 1                  & Daily  \\ \hline
\multirow{3}{*}{\textbf{METEOGALICIA}} & Solar irradiation                              & 1                     & 1                  & Daily  \\
                                       & Sunshine hours                                 & 1                     & 1                  & Daily  \\
                                       & Insolation                                     & 1                     & 1                  & Daily  \\ \hline
\textbf{IEO}                           & Upwelling index                                & 1                     & 1                  & Daily  \\ \hline
\textbf{-}                             & Seasonality                                    & 1                     & 1                  & Daily  \\ \hline
\end{tabular}%
}
\caption{Table of variables (2004-2018)}
\label{tab:variables}
\end{table}

The pre-processing of the input data was as follows:

\begin{itemize}
    \item The weekly information on chlorophyll-a is collected in three samples divided by depth bands: mean chlorophyll-a between 0 and 5 metres, between 5 and 10 metres and between 10 and 15 metres. Since the presence of toxicity in mussel from any part of the culture rope means the total closure of the production area, the maximum value between the three depths was chosen.
    \item The count of \textit{Dynophysis acuminata} is a single, weekly value, so information from all available stations was used.
    \item Nutrient data are collected on a weekly basis and there is only a single piece of data per station, so the count from each oceanographic station was used.
    \item Environmental values, such as temperature and oxygen, were averaged to unify the information into a single measurement since the data are originally irregular measurements at depths between 0 and 25 metres. Only values up to 12 metres were used for averaging, as this is the length of the mussel ropes. In addition, with the temperature and salinity values, a differential was made between the mean of the first 6 metres and that of the following 6 metres, in order to be able to detect the presence of stratifications, both thermoclines and haloclines.
    \item The sun data, such as hours of incidence, insolation and irradiation, come from the Meteogalicia weather station, so the data are daily and common for the whole estuary. In order to simplify the input parameters, the weekly average of each of the parameters was calculated.
    \item The upwelling index data are calculated on a daily basis over four time periods: 00:00 hours, 06:00 hours, 12:00 hours and 18:00 hours. In order to simplify the data, the weekly average value was used, thus estimating the predominant value throughout the week.
    \item To simplify the seasonality into a single value, the date of sampling was transformed, using only the number of the week of the year.
    \item The specific value of toxins in mussel flesh is a value for which no regular records are covering the whole casuistry in a robust way. Instead, it was concluded that it was possible to classify the status of production areas according to whether the growing area was closed or not. These closures are applied in case the level of toxicity in the mussel flesh exceeds the legal threshold. This information could be obtained by analysing INTECMAR's historical record of closures (\cite{intecmar2}).
\end{itemize}

The processing of the 15 years' data resulted in an input dataset of 783 samples. Each of the samples consists of 76 input features. For a more detailed analysis of the input parameters, see table \ref{tab:inputs}.

\begin{table}[h].
  \centering
  \includegraphics[width=1\textwidth]{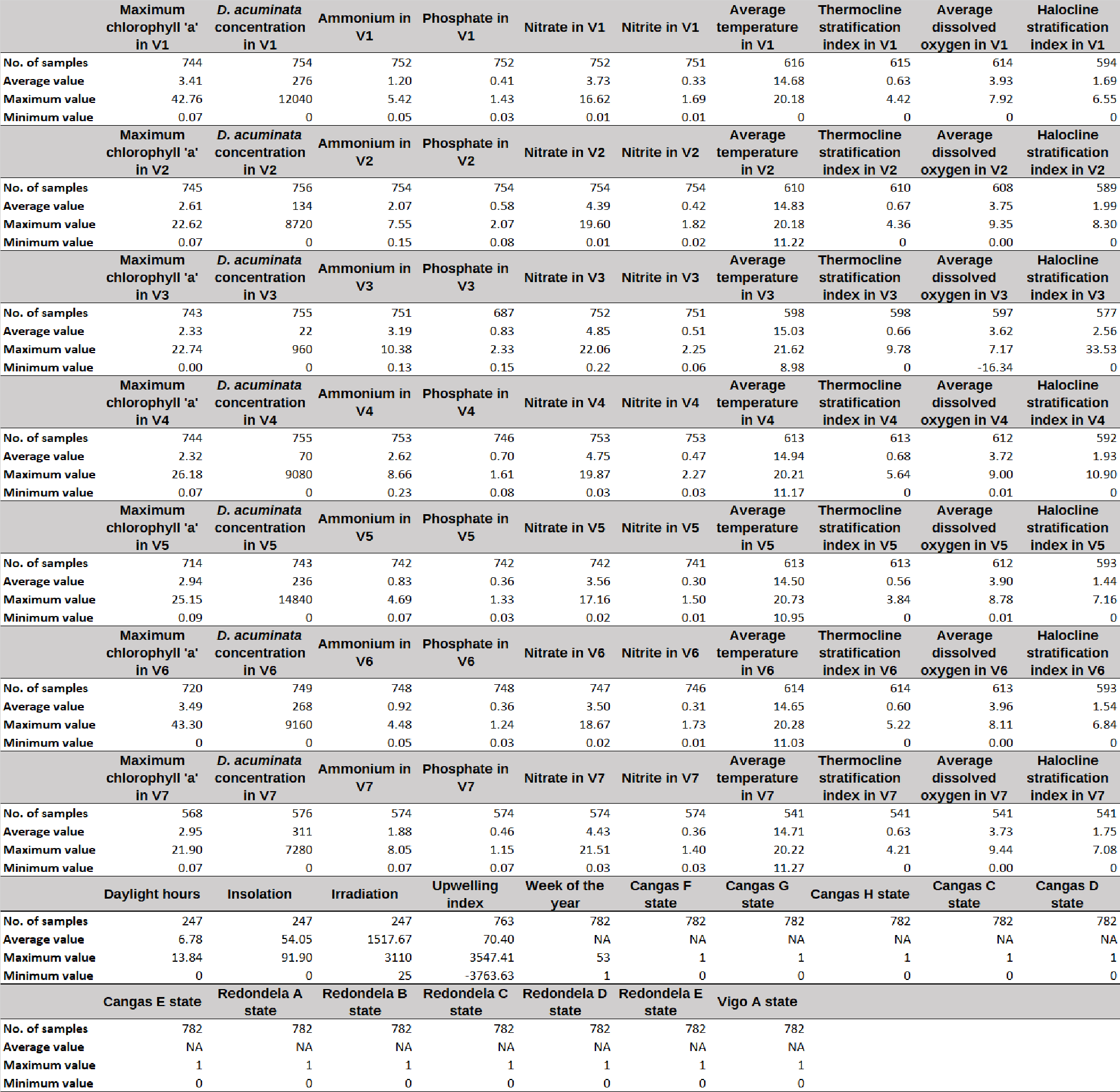}
  \caption{Descriptive analysis of input features}
  \label{tab:inputs}
\end{table}

\begin{figure}[h].
  \centering
  \includegraphics[width=1\textwidth]{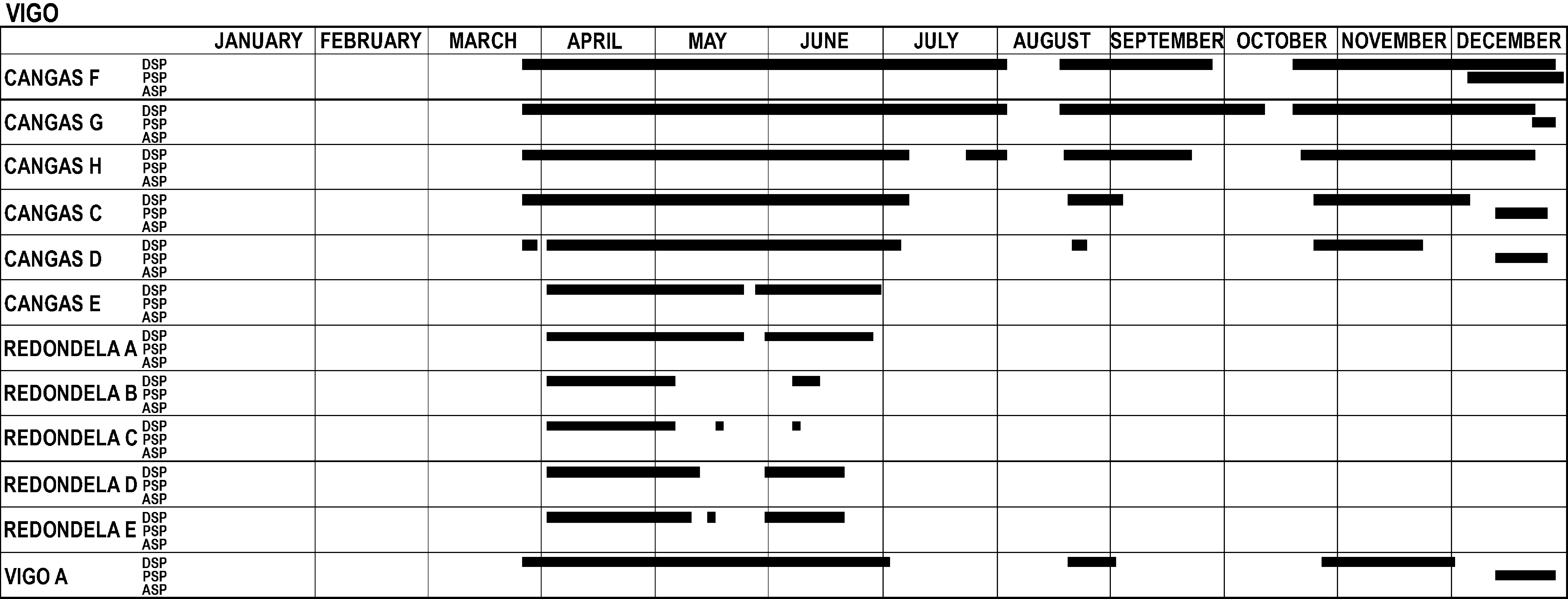}
  \caption{Distribution of closure episodes caused by HAB in mussel production areas in the Vigo Estuary (2016). Source: \url{http://www.intecmar.gal/Informacion/biotoxinas/Evolucion/DiagramaBateas.aspx}}
  \label{fig:ejemplo2016_hd}
\end{figure}

This dataset had incomplete samples with missing data for some of the features, so it was necessary to eliminate those rows with such inconsistencies in their data. These samples with missing values were referred to as null samples. After this filtering, a resulting dataset of 175 samples was left. The distribution in the labelling of the samples can be seen in the table \ref{tab:dist}. As can be seen in this table, toxicity episodes are more common in the crop areas located in the outer part of the estuary, while their frequency decreases towards the inner parts of the estuary. Figure \ref{fig:ejemplo2016_hd} shows the behaviour of the HABs that occurred in 2016. An input dataset was created for each of the twelve crop zones; these matrices share 75 of the 76 input features, with the exception of the Friday opening or closing status of the zone to be estimated.

\begin{table}[h]
\centering
\resizebox{\textwidth}{!}{%
\begin{tabular}{@{}c|cccccccccccc@{}}
\toprule
 &
  \textbf{CangasF} &
  \textbf{CangasG} &
  \textbf{CangasH} &
  \textbf{CangasC} &
  \textbf{CangasD} &
  \textbf{CangasE} &
  \textbf{RedondelaA} &
  \textbf{RedondelaB} &
  \textbf{RedondelaC} &
  \textbf{RedondelaD} &
  \textbf{RedondelaE} &
  \textbf{VigoA} \\ \midrule
\textbf{Samples} &
  783 &
  783 &
  783 &
  783 &
  783 &
  783 &
  783 &
  783 &
  783 &
  783 &
  783 &
  783 \\ \midrule
\textbf{Openings} &
  \begin{tabular}[c]{@{}c@{}}52\%\\ (405)\end{tabular} &
  \begin{tabular}[c]{@{}c@{}}54\%\\ (420)\end{tabular} &
  \begin{tabular}[c]{@{}c@{}}59\%\\ (459)\end{tabular} &
  \begin{tabular}[c]{@{}c@{}}71\%\\ (559)\end{tabular} &
  \begin{tabular}[c]{@{}c@{}}71\%\\ (555)\end{tabular} &
  \begin{tabular}[c]{@{}c@{}}84\%\\ (657)\end{tabular} &
  \begin{tabular}[c]{@{}c@{}}90\%\\ (704)\end{tabular} &
  \begin{tabular}[c]{@{}c@{}}95\%\\ (745)\end{tabular} &
  \begin{tabular}[c]{@{}c@{}}96\%\\ (749)\end{tabular} &
  \begin{tabular}[c]{@{}c@{}}94\%\\ (737)\end{tabular} &
  \begin{tabular}[c]{@{}c@{}}90\%\\ (703)\end{tabular} &
  \begin{tabular}[c]{@{}c@{}}76\%\\ (597)\end{tabular} \\ \midrule
\textbf{Closures} &
  \begin{tabular}[c]{@{}c@{}}48\%\\ (378)\end{tabular} &
  \begin{tabular}[c]{@{}c@{}}46\%\\ (363)\end{tabular} &
  \begin{tabular}[c]{@{}c@{}}41\%\\ (324)\end{tabular} &
  \begin{tabular}[c]{@{}c@{}}29\%\\ (224)\end{tabular} &
  \begin{tabular}[c]{@{}c@{}}29\%\\ (228)\end{tabular} &
  \begin{tabular}[c]{@{}c@{}}16\%\\ (126)\end{tabular} &
  \begin{tabular}[c]{@{}c@{}}10\%\\ (79)\end{tabular} &
  \begin{tabular}[c]{@{}c@{}}5\%\\ (38)\end{tabular} &
  \begin{tabular}[c]{@{}c@{}}4\%\\ (34)\end{tabular} &
  \begin{tabular}[c]{@{}c@{}}6\%\\ (46)\end{tabular} &
  \begin{tabular}[c]{@{}c@{}}10\%\\ (80)\end{tabular} &
  \begin{tabular}[c]{@{}c@{}}24\%\\ (186)\end{tabular} \\ \midrule
\textbf{\begin{tabular}[c]{@{}c@{}}Non-null \\ samples \end{tabular}} &
  175 &
  175 &
  175 &
  175 &
  175 &
  175 &
  175 &
  175 &
  175 &
  175 &
  175 &
  175 \\ \midrule
\textbf{\begin{tabular}[c]{@{}c@{}}Non-null\\ openings\end{tabular}} &
  \begin{tabular}[c]{@{}c@{}}45\%\\ (78)\end{tabular} &
  \begin{tabular}[c]{@{}c@{}}46\%\\ (81)\end{tabular} &
  \begin{tabular}[c]{@{}c@{}}54\%\\ (95)\end{tabular} &
  \begin{tabular}[c]{@{}c@{}}65\%\\ (113)\end{tabular} &
  \begin{tabular}[c]{@{}c@{}}66\%\\ (115)\end{tabular} &
  \begin{tabular}[c]{@{}c@{}}80\%\\ (140)\end{tabular} &
  \begin{tabular}[c]{@{}c@{}}82\%\\ (143)\end{tabular} &
  \begin{tabular}[c]{@{}c@{}}90\%\\ (158)\end{tabular} &
  \begin{tabular}[c]{@{}c@{}}93\%\\ (162)\end{tabular} &
  \begin{tabular}[c]{@{}c@{}}89\%\\ (155)\end{tabular} &
  \begin{tabular}[c]{@{}c@{}}86\%\\ (151)\end{tabular} &
  \begin{tabular}[c]{@{}c@{}}68\%\\ (119)\end{tabular} \\ \midrule
\textbf{\begin{tabular}[c]{@{}c@{}}Non-null\\ closures\end{tabular}} &
  \begin{tabular}[c]{@{}c@{}}55\%\\ (97)\end{tabular} &
  \begin{tabular}[c]{@{}c@{}}54\%\\ (94)\end{tabular} &
  \begin{tabular}[c]{@{}c@{}}46\%\\ (80)\end{tabular} &
  \begin{tabular}[c]{@{}c@{}}35\%\\ (62)\end{tabular} &
  \begin{tabular}[c]{@{}c@{}}34\%\\ (60)\end{tabular} &
  \begin{tabular}[c]{@{}c@{}}20\%\\ (35)\end{tabular} &
  \begin{tabular}[c]{@{}c@{}}18\%\\ (32)\end{tabular} &
  \begin{tabular}[c]{@{}c@{}}10\%\\ (17)\end{tabular} &
  \begin{tabular}[c]{@{}c@{}}7\%\\ (13)\end{tabular} &
  \begin{tabular}[c]{@{}c@{}}11\%\\ (20)\end{tabular} &
  \begin{tabular}[c]{@{}c@{}}14\%\\ (24)\end{tabular} &
  \begin{tabular}[c]{@{}c@{}}32\%\\ (56)\end{tabular} \\ \bottomrule
\end{tabular}%
}
\caption{Distribution of the status of production areas. Non-null sample values refer to samples in which there are no missing values.}
\label{tab:dist}
\end{table}

\subsection{Machine Learning Models}

Based on previous literature, a total of 6 machine learning techniques have been considered: Artificial Neural Networks, Support Vector Machines, XGBoost, k-Nearest Neighbor, Random Forest and Na\"ive Bayes. These techniques will be tested in order to check which method is the most suitable for the approach proposed in this study. These well-known techniques will be briefly presented below.

\subsubsection{Artificial Neural Networks}

Artificial neural networks (ANNs) are massively parallel interconnected networks of simple (usually adaptive) elements and hierarchical organisation. Artificial neural networks are part of a data analysis technique that, compared to their more rigid and complicated alternatives, offers greater flexibility in processing large volumes of multivariate, non-linear data (\cite{white1992artificial}).

\subsubsection{Vector Support Machines}

The classification-regression method Support Vector Machines (SVM) was first proposed by Cortes and Vapnik in 1995 (\cite{cortes1995support}), within the field of computer science. The machine conceptually implements the idea that input vectors are mapped non-linearly into a very high-dimensional feature space. A linear decision surface is constructed in this feature space. The special properties of the decision surface guarantee a high generalisation ability of the learning machine.

\subsubsection{XGBoost}

XGBoost or Extreme Gradient Boosting is an extensible, state-of-the-art application of gradient boosting machines and has been shown to overcome the limits of the computational power of Boosted tree algorithms. Boosting is an ensemble technique in which new models are added to correct errors in existing models. Models are added recursively until no noticeable improvement is found. Gradient boosting is an algorithm in which new models are created to predict the residuals of previous models and then added together to produce a final prediction. It uses a gradient descent algorithm to minimise losses when adding new models (\cite{friedman2001greedy}).

\subsubsection{k-Nearest Neighbour}

The \textit{k-Nearest Neighbor} (kNN) classifier is an unsupervised machine learning technique for classifying unlabeled observations by assigning them to the class of the most similar labelled examples. The features of the observations are collected for both training and test dataset. The most commonly used metric in the calculations is the Euclidean distance. Another concept is the parameter $k$, which decides how many neighbours will be chosen for the kNN algorithm. The appropriate choice of $k$ has a significant impact on the diagnostic performance of the kNN algorithm (\cite{lantz2013machine}).

\subsubsection{Random Forest}
\textit{Random Forest} is an ensemble method, which builds many decision trees that will be used to rank a new instance based on the majority vote. Each node of the decision tree uses a subset of features randomly selected from the original set of features. In addition, each tree uses a different bootstrap data sample, in the same way as bagging. Bagging methods are almost always more accurate than single classifiers. On the other hand, boosting methods can be more accurate than bagging methods but are very sensitive to noise. Random Forest is more robust to noise than boosting methods; performs as well as boosting and sometimes better; and does not overfit (\cite{segal2004machine}).

\subsubsection{Na\"ive Bayes}

Today, the \textit{Na\"ive Bayes} classifier is used in many applications due to its simple but powerful principle of (\cite{lewis1998naive}) accuracy. Bayes' theorem finds the probability of an event occurring given the probability that another event has already occurred. However, this classifier does not take into account the number of occurrences, which is a potentially useful source of additional information. They are called ``na\"ive'' because the algorithm assumes that all terms occur independently of each other.

\subsection{Performance mesuares}

For the analysis of the trained models and their subsequent comparison, 6 statistics were taken into account that were considered relevant when assessing the results (average accuracy, average sensitivity, average kappa coefficient, minimum accuracy, minimum sensitivity and minimum kappa coefficient). In the confusion matrix used to calculate the statistics, closures were defined as positive and openings as negative. Thus, True Positives ($TP$) correspond to those closures correctly classified as closures, True Negatives ($TN$) identify openings classified as such, False Positives ($FP$) represent those openings wrongly classified as closures and, finally, False Negatives ($FN$) are those closures that have been classified as openings.\\ 
Calculated according to Eq. \ref{eq:accuracy} accuracy estimates how correctly a binary classification test identifies or excludes a condition. As this is a binary classification paper, this parameter is considered relevant.

\begin{equation}
   \frac{TP+TN}{TP+FP+FN+TN}
   \label{eq:accuracy}
\end{equation}

Not performing a closure when the toxin is present in the mussel poses a higher risk, prioritising the human factor over the economic one. Sensitivity (Eq. \ref{eq:sensibility}) prioritises avoiding misclassifying closures as openings. Sensitivity was therefore the benchmark statistic in this study.

\begin{equation}
    \frac{TP}{TP+FN}
    \label{eq:sensibility}
\end{equation}

Cohen's kappa coefficient, calculated according to Eq. \ref{eq:kappa}, is a statistical measure that adjusts for the effect of chance on the proportion of observed agreement between two experts. In this equation, $Pr(a)$ represents the relative observed agreement between the observers, while $Pr(e)$ is the hypothetical probability of agreement by chance. In this study, the model outputs were compared with the labelling performed by the experts to analyse the effect of chance on the models.

\begin{equation}
    K=\frac{Pr(a)-Pr(e)}{1-Pr(e)}
    \label{eq:kappa}
\end{equation}

The criteria taken into account when selecting the best models were the values explained above (accuracy, sensitivity and kappa coefficient), as well as the number of features used to make the prediction. A smaller number of input variables would make it easier to make predictions, even on days when certain data are missing. Sensitivity is the most important factor to be taken into account due to the absolute priority of minimising false negatives (as they pose a risk to public health).

\subsection{Experimentation setup}

By using the strategy of \textit{K-folds} strategy, specifically \textit{10-fold}, yields 10 values of each statistic. The \textit{K-fold cross-validation} procedure randomly divides a dataset into $k$ disjoint blocks of approximately equal size, and each block is in turn used to test the model induced from the other $k-1$ blocks by a classification algorithm. The performance of the classification algorithm is evaluated by the average of the $k$-precisions resulting from the cross-validation of $k$-blocks. This method avoids choosing models with good averages but which perform poorly on certain training blocks, thus ensuring the robustness of the models. The minimum values of the statistics explained above are also taken into account.\\
Significance analysis was deemed necessary to ensure the robustness of the classification. First, a normality analysis was performed, to ensure that a parametric test can be performed (\cite{sheskin2003handbook}). When the sample size is at most 50, normality can be tested with the Shapiro-Wilk test test. The Anderson-Darling statistic measures how well the data follow a specific distribution. For a particular data set and distribution, the better the distribution fits the data, the lower this statistic will be. Both the Shapiro-Wilk test and the Anderson-Darling test showed that the sensitivity data for all areas are normal. ANOVA analysis allows multiple means to be compared by studying variances. This was followed by pairwise comparison, in the specific case of this project, with the Tukey-Kramer test. The significance was estimated according to Copenhaver-Holland (\cite{copenhaver1988computation}).\\
For this study two sets of features were used: one with all 76 input features and another one where the most redundant features were filtered out. In order to do this, a correlation analysis was carried out between the features, and those with a correlation of more than 90\% between them were eliminated. This was an empirical approach in which preliminary tests were carried out to eliminate only those variables that really had a very close relationship and leave it to a more purely objective process such as a ranking system to use or assign importance to each. Through this process, influential factors have been sought in less common variables. In this second approach, the 76 input features were reduced to 50.\\
Then, in each approach, starting from the raw data, a feature selection process was conducted. This has several advantages. Firstly, we make our model easier to interpret. Secondly, we can reduce the variance of the model and thus the overfitting. Finally, we can reduce the computational cost (and time) of training a model. To carry out the feature selection process, the features were ordered using a ranking process. Two ranking techniques were used for this process:

\begin{itemize}
\item Applying a filtering method such as correlation with the variable to be forecast. Using the statistical value to rank order the features, three sets of tests were proposed: one with 25\% of the best ranking features, one with 50\% and the last one with 75\%.
\item Use of an embedded method such as the Random Forest algorithm. The tree-based strategies used by Random Forest are naturally ranked according to how they improve node purity. This means a decrease in impurity over all trees (called Gini impurity). Nodes with the highest decrease in impurity occur at the beginning of the trees, while notes with the lowest decrease in impurity occur at the end of the trees. Thus, by pruning the trees below a particular node, we can create a subset of the most important features. After applying this ranking, three sets of tests were proposed: one with 25\% of the best ranking features, one with 50\% and the last one with 75\%.
\end{itemize}

Different experiments have been defined based on the application of one, both or none of the ranking methods mentioned above. To ensure the reliability of the results, the tests were carried out with a cross-validation strategy of \textit{10-fold}. In order to determine the configuration of the best performing models, a grid search was performed and the parameter values of the models used in the training were adjusted as shown in the table \ref{tab:configuracion}.

\begin{table}[h]
\centering
\resizebox{1\textwidth}{!}{%
\begin{tabular}{|lc|}
\hline
\rowcolor[HTML]{C0C0C0} 
\textbf{General Settings}     & \multicolumn{1}{l|}{\cellcolor[HTML]{C0C0C0}} \\ \hline
Validation strategy  & 10-fold cross-validation       \\ 
Data normalisation    & Yes                             \\ \hline
\rowcolor[HTML]{EFEFEF} 
\textbf{Artificial Neural Networks}     & \multicolumn{1}{l|}{\cellcolor[HTML]{EFEFEF}} \\ \hline
Number of input neurons              & Number of influencing factors              \\ 
Output neurons                       & 1                        \\ 
Number of hidden layers                  & 1 and 2                    \\ 
Number of neurons in a one hidden layer network & 2, 8 and 14                \\ 
Number of neurons in a two hidden layers network & {[}10,10{]} and {[}10,20{]}                     \\ 
Activation function output layer        & Sigmoid                 \\ 
Hidden layers activation function      & Relu                     \\ 
Optimizer                              & Adam                     \\ 
Learning rate                            & 0,001                    \\ 
Loss function                       & Binary crossentropy      \\ 
Batch size                              & 5                        \\ 
Number of epochs                         & 10                       \\ 
Class weighting                & Yes                       \\ \hline
\rowcolor[HTML]{EFEFEF} 
\textbf{Support Vector Machines} & \multicolumn{1}{l|}{\cellcolor[HTML]{EFEFEF}} \\ \hline
Kernel type            & Lineal, Gaussian and Polynomial \\ 
C value                & 1                              \\ 
Gamma value (gaussian kernel)        & 0.2, 0.3, 0.4, 0.5, 0.6, 0.7 and 0.8            \\ 
Grade (polynomial kernel) & 2                              \\ \hline
\rowcolor[HTML]{EFEFEF} 
\textbf{XGBoost}         & \multicolumn{1}{l|}{\cellcolor[HTML]{EFEFEF}} \\ \hline
\rowcolor[HTML]{EFEFEF} 
\textbf{Gbtree}          &                                               \\ 
Max depth       & 6                                             \\ 
Learning rate            & 0,3                                           \\ 
\rowcolor[HTML]{EFEFEF} 
\textbf{Dart}            &                                               \\ 
Sample type              & uniform                                       \\ 
Normalise type           & forest                                        \\
\rowcolor[HTML]{EFEFEF} 
\textbf{Gblinear}        &                                               \\ 
Updater                  & coord\_descent                                \\ \hline
\rowcolor[HTML]{EFEFEF} 
\textbf{k-Nearest   Neighbor} & \multicolumn{1}{l|}{\cellcolor[HTML]{EFEFEF}} \\ \hline
k value                    & {1, 2, 3, 4, 5, 6, 7, 8, 9 and 10}                                    \\ \hline
\rowcolor[HTML]{EFEFEF} 
\textbf{Random   Forest} & \multicolumn{1}{l|}{\cellcolor[HTML]{EFEFEF}} \\ \hline
Number of trees        & 100, 500, 1000, 1500 and 2000                                           \\ \hline
\rowcolor[HTML]{EFEFEF} 
\textbf{Na\"ive Bayes}     & \multicolumn{1}{l|}{\cellcolor[HTML]{EFEFEF}} \\ \hline
Algorithm                & Gaussian, Multinomial, Complement and Bernoulli \\ \hline
\end{tabular}%
}
\caption{Summary table of the models parameter values used in the grid search.}
\label{tab:configuracion}
\end{table}

\section{Results}
\label{sec:r}

\begin{figure}[h].
  \centering
  \includegraphics[width=1\textwidth]{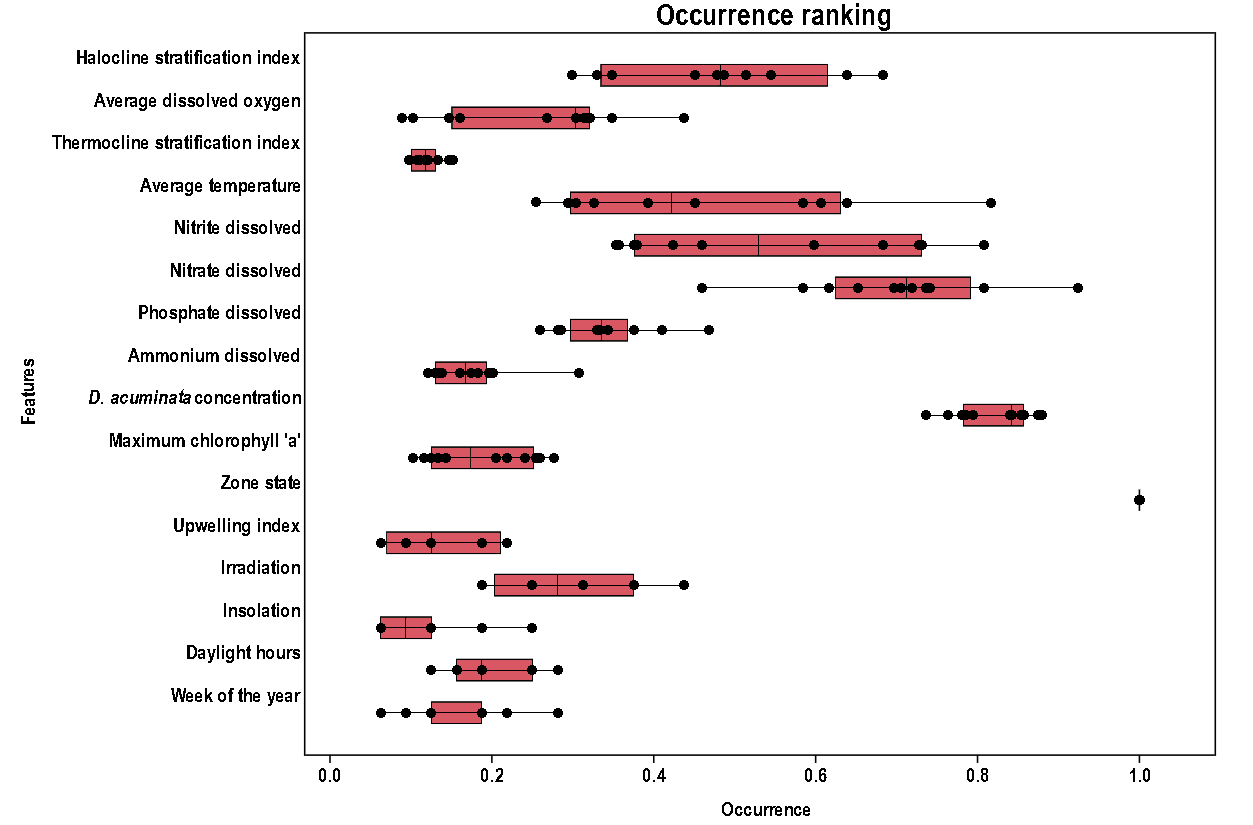}
  \caption{Summary table with the occurrence of the features after the feature selection processes. Where each point represents the likelihood of a variable being selected as an input feature for a particular production area.}
  \label{fig:ranking}
\end{figure}

During the feature selection process, the combined Pearson Correlation and Random Forest techniques were applied. Thanks to this, it was possible to extract the importance that these methods give to the features for the classification process. Figure \ref{fig:ranking} shows a summary of the behaviour of these methods throughout the production zones, reflecting the percentage of persistence of each variable after the selection processes. It can be seen that the state of the production zone in the week before the prediction day is the most important characteristic, followed by the concentration of \textit{D. accuminata} and the concentration of dissolved nutrients such as nitrate and nitrite. For each of the production zones, a more detailed overview of the feature selection process can be found in tables \ref{fig:rank_cangasF}, \ref{fig:rank_cangasG}, \ref{fig:rank_cangasH}, \ref{fig:rank_cangasC}, \ref{fig:rank_cangasD}, \ref{fig:rank_cangasE}, \ref{fig:rank_redondelaA}, \ref{fig:rank_redondelaB}, \ref{fig:rank_redondelaC}, \ref{fig:rank_redondelaD}, \ref{fig:rank_redondelaE} and \ref{fig:rank_vigoA}. These tables show how the data collected at each oceanographic station have a different effect on nearby areas. This is due to how marine currents affect the estuary and how certain stations gain importance over others concerning each production area.

\begin{figure}[h].
  \centering
  \includegraphics[width=1\textwidth]{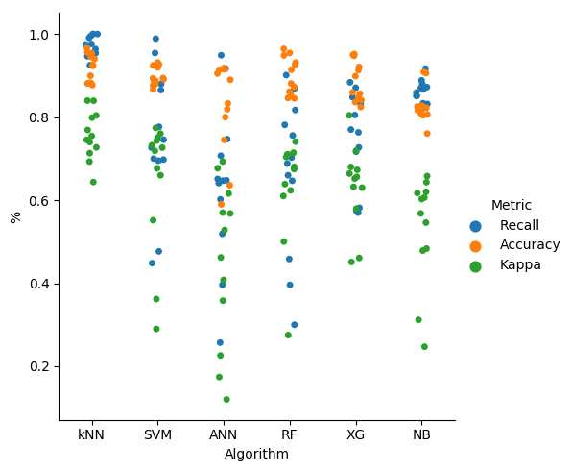}
  \caption{Combination of recall, accuracy and kappa in the different production zones for each algorithm. The average values for each metric across all folds are shown. In each case, the best performing configurations are represented.}
  \label{fig:models}
\end{figure}

By applying each of the 6 machine learning techniques to the 12 production zones independently, it has been possible to observe the comparative solutions offered by each of these methodologies. In figure \ref{fig:models} the values of sensitivity, accuracy and kappa obtained by the best models trained with each algorithm and for each production zone can be seen. Algorithms such as kNN or NB obtain more stable results for all the zones, while algorithms such as SVM, RF and XG, although they show certain stability in the values of accuracy, show great variability in the sensitivity values depending on the production zone. The ANN algorithm is presented as the algorithm with the greatest variability in its results.

\begin{figure}[h].
  \centering
  \includegraphics[width=1\textwidth]{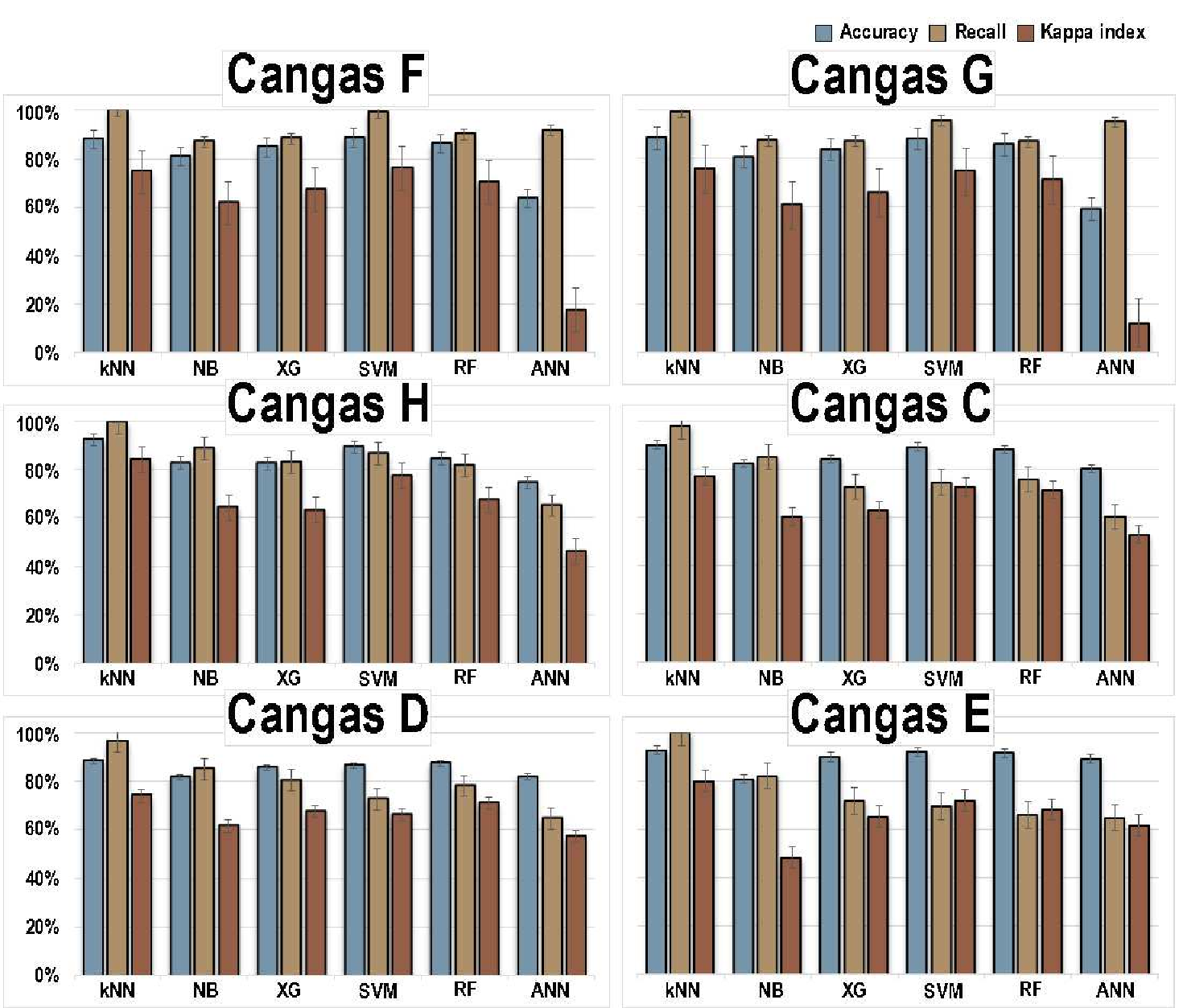}
  \caption{Combination of recall, accuracy and kappa in the production zones for each algorithm. The average values for each metric across all folds are shown. In each case, the best performing configurations are represented. 1/2}
  \label{fig:conjunto1}
\end{figure}

\begin{figure}[h].
  \centering
  \includegraphics[width=1\textwidth]{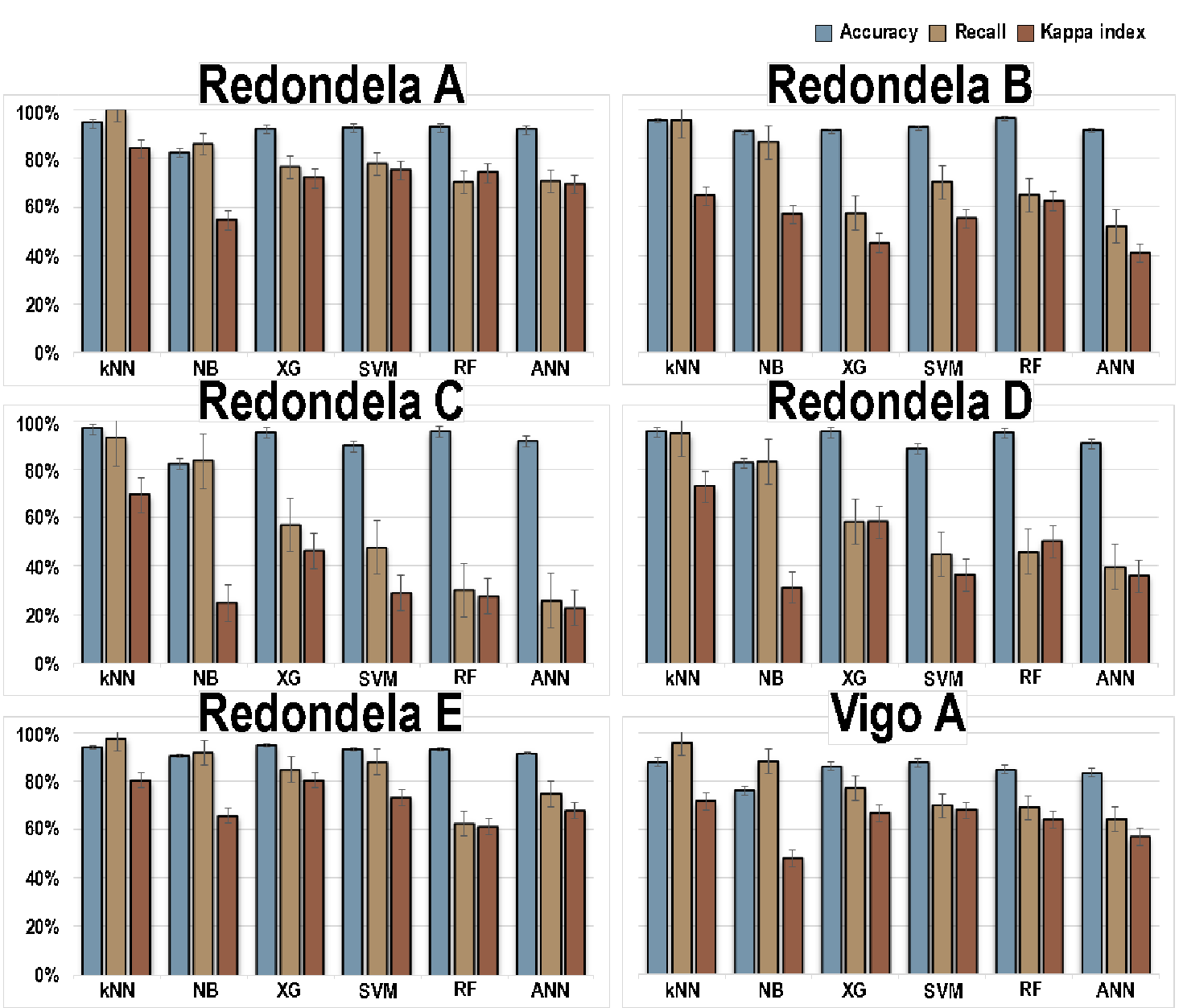}
  \caption{Combination of recall, accuracy and kappa in the production zones for each algorithm. The average values for each metric across all folds are shown. In each case, the best performing configurations are represented. 2/2}
  \label{fig:conjunto2}
\end{figure}

For a detailed analysis of how the algorithms behave in each of the production zones, please refer to the figures \ref{fig:conjunto1} and \ref{fig:conjunto2}. In these graphs it can be seen that the models perform better in the production areas of Cangas F, Cangas G, Cangas H and Redondela A. While the areas where the models have more difficulties in making predictions are: Redondela B, Redondela C and Redondela D.

The models defined as the best in each of the production zones during the first approach are shown on table \ref{tab:tabla_1} and those of the second approach on table \ref{tab:tabla_2}. These tables show the sensitivity, accuracy and kappa values. When applying the ten-folds strategy, it is necessary to show the results as the tuple of mean value and standard deviation of the values obtained in each fold.

\begin{table}[h]
\centering
\resizebox{\textwidth}{!}{%
\begin{tabular}{@{}ccccccccccc@{}}
\toprule
\rowcolor[HTML]{EFEFEF} 
\multicolumn{11}{c}{\cellcolor[HTML]{EFEFEF}\textbf{Approach 1}} \\ \midrule
\rowcolor[HTML]{EFEFEF} 
\multicolumn{5}{|c|}{\cellcolor[HTML]{EFEFEF}} & \multicolumn{2}{c|}{\cellcolor[HTML]{EFEFEF}\textbf{Recall}} & \multicolumn{2}{c|}{\cellcolor[HTML]{EFEFEF}\textbf{Accuracy}} & \multicolumn{2}{c|}{\cellcolor[HTML]{EFEFEF}\textbf{Kappa}} \\ \midrule
\rowcolor[HTML]{EFEFEF} 
\multicolumn{1}{|c|}{\cellcolor[HTML]{EFEFEF}\textbf{Production}} & \multicolumn{1}{c|}{\cellcolor[HTML]{EFEFEF}\textbf{Corelation}} & \multicolumn{1}{c|}{\cellcolor[HTML]{EFEFEF}\textbf{Random Forest}} & \multicolumn{1}{c|}{\cellcolor[HTML]{EFEFEF}} & \multicolumn{1}{c|}{\cellcolor[HTML]{EFEFEF}\textbf{Number of}} & \multicolumn{1}{c|}{\cellcolor[HTML]{EFEFEF}} & \multicolumn{1}{c|}{\cellcolor[HTML]{EFEFEF}} & \multicolumn{1}{c|}{\cellcolor[HTML]{EFEFEF}} & \multicolumn{1}{c|}{\cellcolor[HTML]{EFEFEF}} & \multicolumn{1}{c|}{\cellcolor[HTML]{EFEFEF}} & \multicolumn{1}{c|}{\cellcolor[HTML]{EFEFEF}} \\
\rowcolor[HTML]{EFEFEF} 
\multicolumn{1}{|c|}{\cellcolor[HTML]{EFEFEF}\textbf{zone}} & \multicolumn{1}{c|}{\cellcolor[HTML]{EFEFEF}\textbf{filter cuartile}} & \multicolumn{1}{c|}{\cellcolor[HTML]{EFEFEF}\textbf{filter cuartile}} & \multicolumn{1}{c|}{\multirow{-2}{*}{\cellcolor[HTML]{EFEFEF}\textbf{Algorithm}}} & \multicolumn{1}{c|}{\cellcolor[HTML]{EFEFEF}\textbf{neighbors}} & \multicolumn{1}{c|}{\multirow{-2}{*}{\cellcolor[HTML]{EFEFEF}\textbf{$\mu$}}} & \multicolumn{1}{c|}{\multirow{-2}{*}{\cellcolor[HTML]{EFEFEF}\textbf{$\sigma$}}} & \multicolumn{1}{c|}{\multirow{-2}{*}{\cellcolor[HTML]{EFEFEF}\textbf{$\mu$}}} & \multicolumn{1}{c|}{\multirow{-2}{*}{\cellcolor[HTML]{EFEFEF}\textbf{$\sigma$}}} & \multicolumn{1}{c|}{\multirow{-2}{*}{\cellcolor[HTML]{EFEFEF}\textbf{$\mu$}}} & \multicolumn{1}{c|}{\multirow{-2}{*}{\cellcolor[HTML]{EFEFEF}\textbf{$\sigma$}}} \\ \midrule
\multicolumn{1}{|c|}{\textbf{Cangas F}} & \multicolumn{1}{c|}{-} & \multicolumn{1}{c|}{50} & \multicolumn{1}{c|}{kNN} & \multicolumn{1}{c|}{2} & \multicolumn{1}{c|}{100,00\%} & \multicolumn{1}{c|}{0,00\%} & \multicolumn{1}{c|}{91,38\%} & \multicolumn{1}{c|}{6,37\%} & \multicolumn{1}{c|}{0,79} & \multicolumn{1}{c|}{0,14} \\ \midrule
\multicolumn{1}{|c|}{\textbf{Cangas G}} & \multicolumn{1}{c|}{25} & \multicolumn{1}{c|}{25} & \multicolumn{1}{c|}{kNN} & \multicolumn{1}{c|}{4} & \multicolumn{1}{c|}{99,17\%} & \multicolumn{1}{c|}{2,50\%} & \multicolumn{1}{c|}{88,50\%} & \multicolumn{1}{c|}{7,23\%} & \multicolumn{1}{c|}{0,75} & \multicolumn{1}{c|}{0,13} \\ \midrule
\multicolumn{1}{|c|}{\textbf{Cangas H}} & \multicolumn{1}{c|}{75} & \multicolumn{1}{c|}{75} & \multicolumn{1}{c|}{kNN} & \multicolumn{1}{c|}{2} & \multicolumn{1}{c|}{99,50\%} & \multicolumn{1}{c|}{1,50\%} & \multicolumn{1}{c|}{91,98\%} & \multicolumn{1}{c|}{3,56\%} & \multicolumn{1}{c|}{0,83} & \multicolumn{1}{c|}{0,08} \\ \midrule
\multicolumn{1}{|c|}{\textbf{Cangas C}} & \multicolumn{1}{c|}{50} & \multicolumn{1}{c|}{75} & \multicolumn{1}{c|}{kNN} & \multicolumn{1}{c|}{2} & \multicolumn{1}{c|}{97,61\%} & \multicolumn{1}{c|}{2,97\%} & \multicolumn{1}{c|}{89,23\%} & \multicolumn{1}{c|}{3,64\%} & \multicolumn{1}{c|}{0,76} & \multicolumn{1}{c|}{0,09} \\ \midrule
\multicolumn{1}{|c|}{\textbf{Cangas D}} & \multicolumn{1}{c|}{-} & \multicolumn{1}{c|}{-} & \multicolumn{1}{c|}{kNN} & \multicolumn{1}{c|}{2} & \multicolumn{1}{c|}{96,39\%} & \multicolumn{1}{c|}{7,86\%} & \multicolumn{1}{c|}{89,23\%} & \multicolumn{1}{c|}{6,79\%} & \multicolumn{1}{c|}{0,76} & \multicolumn{1}{c|}{0,14} \\ \midrule
\multicolumn{1}{|c|}{\textbf{Cangas E}} & \multicolumn{1}{c|}{-} & \multicolumn{1}{c|}{-} & \multicolumn{1}{c|}{kNN} & \multicolumn{1}{c|}{2} & \multicolumn{1}{c|}{100,00\%} & \multicolumn{1}{c|}{0,00\%} & \multicolumn{1}{c|}{92,61\%} & \multicolumn{1}{c|}{5,02\%} & \multicolumn{1}{c|}{0,80} & \multicolumn{1}{c|}{0,12} \\ \midrule
\multicolumn{1}{|c|}{\textbf{Vigo A}} & \multicolumn{1}{c|}{50} & \multicolumn{1}{c|}{75} & \multicolumn{1}{c|}{kNN} & \multicolumn{1}{c|}{2} & \multicolumn{1}{c|}{96,32\%} & \multicolumn{1}{c|}{3,83\%} & \multicolumn{1}{c|}{88,70\%} & \multicolumn{1}{c|}{4,01\%} & \multicolumn{1}{c|}{0,73} & \multicolumn{1}{c|}{0,09} \\ \midrule
\multicolumn{1}{|c|}{\textbf{Redondela A}} & \multicolumn{1}{c|}{-} & \multicolumn{1}{c|}{-} & \multicolumn{1}{c|}{kNN} & \multicolumn{1}{c|}{2} & \multicolumn{1}{c|}{100,00\%} & \multicolumn{1}{c|}{0,00\%} & \multicolumn{1}{c|}{93,93\%} & \multicolumn{1}{c|}{3,76\%} & \multicolumn{1}{c|}{0,83} & \multicolumn{1}{c|}{0,11} \\ \midrule
\multicolumn{1}{|c|}{\textbf{Redondela B}} & \multicolumn{1}{c|}{-} & \multicolumn{1}{c|}{-} & \multicolumn{1}{c|}{kNN} & \multicolumn{1}{c|}{2} & \multicolumn{1}{c|}{90,83\%} & \multicolumn{1}{c|}{20,56\%} & \multicolumn{1}{c|}{90,83\%} & \multicolumn{1}{c|}{6,90\%} & \multicolumn{1}{c|}{0,64} & \multicolumn{1}{c|}{0,27} \\ \midrule
\multicolumn{1}{|c|}{\textbf{Redondela C}} & \multicolumn{1}{c|}{50} & \multicolumn{1}{c|}{75} & \multicolumn{1}{c|}{kNN} & \multicolumn{1}{c|}{2} & \multicolumn{1}{c|}{92,50\%} & \multicolumn{1}{c|}{16,01\%} & \multicolumn{1}{c|}{96,42\%} & \multicolumn{1}{c|}{1,92\%} & \multicolumn{1}{c|}{0,69} & \multicolumn{1}{c|}{0,13} \\ \midrule
\multicolumn{1}{|c|}{\textbf{Redondela D}} & \multicolumn{1}{c|}{50} & \multicolumn{1}{c|}{75} & \multicolumn{1}{c|}{kNN} & \multicolumn{1}{c|}{2} & \multicolumn{1}{c|}{93,67\%} & \multicolumn{1}{c|}{12,69\%} & \multicolumn{1}{c|}{93,69\%} & \multicolumn{1}{c|}{3,62\%} & \multicolumn{1}{c|}{0,65} & \multicolumn{1}{c|}{0,19} \\ \midrule
\multicolumn{1}{|c|}{\textbf{Redondela E}} & \multicolumn{1}{c|}{50} & \multicolumn{1}{c|}{75} & \multicolumn{1}{c|}{kNN} & \multicolumn{1}{c|}{2} & \multicolumn{1}{c|}{98,33\%} & \multicolumn{1}{c|}{5,00\%} & \multicolumn{1}{c|}{93,63\%} & \multicolumn{1}{c|}{5,15\%} & \multicolumn{1}{c|}{0,82} & \multicolumn{1}{c|}{0,15} \\ \midrule
\rowcolor[HTML]{EFEFEF} 
\multicolumn{4}{l}{\cellcolor[HTML]{EFEFEF}\textit{\textbf{Average}}} &  & \textit{\textbf{97,03\%}} & \textit{\textbf{6,08\%}} & \textit{\textbf{91,68\%}} & \textit{\textbf{4,83\%}} & \textit{\textbf{0,76}} & \textit{\textbf{0,14}} \\ \bottomrule
\end{tabular}%
}
\caption{Summary table of the first approach with the models defined as the best in each of the production zones.}
\label{tab:tabla_1}
\end{table}

\begin{table}[h]
\centering
\resizebox{\textwidth}{!}{%
\begin{tabular}{@{}ccccccccccc@{}}
\toprule
\rowcolor[HTML]{EFEFEF} 
\multicolumn{11}{c}{\cellcolor[HTML]{EFEFEF}\textbf{Approach 2}} \\ \midrule
\rowcolor[HTML]{EFEFEF} 
\multicolumn{5}{|c|}{\cellcolor[HTML]{EFEFEF}} & \multicolumn{2}{c|}{\cellcolor[HTML]{EFEFEF}\textbf{Recall}} & \multicolumn{2}{c|}{\cellcolor[HTML]{EFEFEF}\textbf{Accuracy}} & \multicolumn{2}{c|}{\cellcolor[HTML]{EFEFEF}\textbf{Kappa}} \\ \midrule
\rowcolor[HTML]{EFEFEF} 
\multicolumn{1}{|c|}{\cellcolor[HTML]{EFEFEF}\textbf{Production}} & \multicolumn{1}{c|}{\cellcolor[HTML]{EFEFEF}\textbf{Corelation}} & \multicolumn{1}{c|}{\cellcolor[HTML]{EFEFEF}\textbf{Random Forest}} & \multicolumn{1}{c|}{\cellcolor[HTML]{EFEFEF}} & \multicolumn{1}{c|}{\cellcolor[HTML]{EFEFEF}\textbf{Number of}} & \multicolumn{1}{c|}{\cellcolor[HTML]{EFEFEF}} & \multicolumn{1}{c|}{\cellcolor[HTML]{EFEFEF}} & \multicolumn{1}{c|}{\cellcolor[HTML]{EFEFEF}} & \multicolumn{1}{c|}{\cellcolor[HTML]{EFEFEF}} & \multicolumn{1}{c|}{\cellcolor[HTML]{EFEFEF}} & \multicolumn{1}{c|}{\cellcolor[HTML]{EFEFEF}} \\
\rowcolor[HTML]{EFEFEF} 
\multicolumn{1}{|c|}{\cellcolor[HTML]{EFEFEF}\textbf{zone}} & \multicolumn{1}{c|}{\cellcolor[HTML]{EFEFEF}\textbf{filter cuartile}} & \multicolumn{1}{c|}{\cellcolor[HTML]{EFEFEF}\textbf{filter cuartile}} & \multicolumn{1}{c|}{\multirow{-2}{*}{\cellcolor[HTML]{EFEFEF}\textbf{Algorithm}}} & \multicolumn{1}{c|}{\cellcolor[HTML]{EFEFEF}\textbf{neighbors}} & \multicolumn{1}{c|}{\multirow{-2}{*}{\cellcolor[HTML]{EFEFEF}\textbf{$\mu$}}} & \multicolumn{1}{c|}{\multirow{-2}{*}{\cellcolor[HTML]{EFEFEF}\textbf{$\sigma$}}} & \multicolumn{1}{c|}{\multirow{-2}{*}{\cellcolor[HTML]{EFEFEF}\textbf{$\mu$}}} & \multicolumn{1}{c|}{\multirow{-2}{*}{\cellcolor[HTML]{EFEFEF}\textbf{$\sigma$}}} & \multicolumn{1}{c|}{\multirow{-2}{*}{\cellcolor[HTML]{EFEFEF}\textbf{$\mu$}}} & \multicolumn{1}{c|}{\multirow{-2}{*}{\cellcolor[HTML]{EFEFEF}\textbf{$\sigma$}}} \\ \midrule
\multicolumn{1}{|c|}{\textbf{Cangas F}} & \multicolumn{1}{c|}{-} & \multicolumn{1}{c|}{-} & \multicolumn{1}{c|}{kNN} & \multicolumn{1}{c|}{2} & \multicolumn{1}{c|}{100,00\%} & \multicolumn{1}{c|}{0,00\%} & \multicolumn{1}{c|}{88,10\%} & \multicolumn{1}{c|}{7,52\%} & \multicolumn{1}{c|}{0,75} & \multicolumn{1}{c|}{0,15} \\ \midrule
\multicolumn{1}{|c|}{\textbf{Cangas G}} & \multicolumn{1}{c|}{25} & \multicolumn{1}{c|}{-} & \multicolumn{1}{c|}{kNN} & \multicolumn{1}{c|}{2} & \multicolumn{1}{c|}{99,09\%} & \multicolumn{1}{c|}{2,73\%} & \multicolumn{1}{c|}{87,73\%} & \multicolumn{1}{c|}{11,46\%} & \multicolumn{1}{c|}{0,74} & \multicolumn{1}{c|}{0,21} \\ \midrule
\multicolumn{1}{|c|}{\textbf{Cangas H}} & \multicolumn{1}{c|}{50} & \multicolumn{1}{c|}{75} & \multicolumn{1}{c|}{kNN} & \multicolumn{1}{c|}{2} & \multicolumn{1}{c|}{99,50\%} & \multicolumn{1}{c|}{1,50\%} & \multicolumn{1}{c|}{92,35\%} & \multicolumn{1}{c|}{3,54\%} & \multicolumn{1}{c|}{0,84} & \multicolumn{1}{c|}{0,08} \\ \midrule
\multicolumn{1}{|c|}{\textbf{Cangas C}} & \multicolumn{1}{c|}{50} & \multicolumn{1}{c|}{75} & \multicolumn{1}{c|}{kNN} & \multicolumn{1}{c|}{2} & \multicolumn{1}{c|}{97,61\%} & \multicolumn{1}{c|}{2,97\%} & \multicolumn{1}{c|}{89,98\%} & \multicolumn{1}{c|}{3,92\%} & \multicolumn{1}{c|}{0,77} & \multicolumn{1}{c|}{0,09} \\ \midrule
\multicolumn{1}{|c|}{\textbf{Cangas D}} & \multicolumn{1}{c|}{75} & \multicolumn{1}{c|}{75} & \multicolumn{1}{c|}{kNN} & \multicolumn{1}{c|}{2} & \multicolumn{1}{c|}{96,39\%} & \multicolumn{1}{c|}{2,58\%} & \multicolumn{1}{c|}{87,35\%} & \multicolumn{1}{c|}{3,91\%} & \multicolumn{1}{c|}{0,73} & \multicolumn{1}{c|}{0,08} \\ \midrule
\multicolumn{1}{|c|}{\textbf{Cangas E}} & \multicolumn{1}{c|}{-} & \multicolumn{1}{c|}{-} & \multicolumn{1}{c|}{kNN} & \multicolumn{1}{c|}{2} & \multicolumn{1}{c|}{100,00\%} & \multicolumn{1}{c|}{0,00\%} & \multicolumn{1}{c|}{92,58\%} & \multicolumn{1}{c|}{5,09\%} & \multicolumn{1}{c|}{0,80} & \multicolumn{1}{c|}{0,14} \\ \midrule
\multicolumn{1}{|c|}{\textbf{Vigo A}} & \multicolumn{1}{c|}{-} & \multicolumn{1}{c|}{25} & \multicolumn{1}{c|}{kNN} & \multicolumn{1}{c|}{2} & \multicolumn{1}{c|}{95,42\%} & \multicolumn{1}{c|}{10,28\%} & \multicolumn{1}{c|}{87,71\%} & \multicolumn{1}{c|}{6,93\%} & \multicolumn{1}{c|}{0,71} & \multicolumn{1}{c|}{0,15} \\ \midrule
\multicolumn{1}{|c|}{\textbf{Redondela A}} & \multicolumn{1}{c|}{-} & \multicolumn{1}{c|}{-} & \multicolumn{1}{c|}{kNN} & \multicolumn{1}{c|}{2} & \multicolumn{1}{c|}{100,00\%} & \multicolumn{1}{c|}{0,00\%} & \multicolumn{1}{c|}{94,48\%} & \multicolumn{1}{c|}{3,36\%} & \multicolumn{1}{c|}{0,84} & \multicolumn{1}{c|}{0,11} \\ \midrule
\multicolumn{1}{|c|}{\textbf{Redondela B}} & \multicolumn{1}{c|}{-} & \multicolumn{1}{c|}{50} & \multicolumn{1}{c|}{kNN} & \multicolumn{1}{c|}{2} & \multicolumn{1}{c|}{95,42\%} & \multicolumn{1}{c|}{10,28\%} & \multicolumn{1}{c|}{95,59\%} & \multicolumn{1}{c|}{3,20\%} & \multicolumn{1}{c|}{0,64} & \multicolumn{1}{c|}{0,24} \\ \midrule
\multicolumn{1}{|c|}{\textbf{Redondela C}} & \multicolumn{1}{c|}{50} & \multicolumn{1}{c|}{75} & \multicolumn{1}{c|}{kNN} & \multicolumn{1}{c|}{2} & \multicolumn{1}{c|}{92,50\%} & \multicolumn{1}{c|}{16,01\%} & \multicolumn{1}{c|}{96,60\%} & \multicolumn{1}{c|}{1,64\%} & \multicolumn{1}{c|}{0,69} & \multicolumn{1}{c|}{0,12} \\ \midrule
\multicolumn{1}{|c|}{\textbf{Redondela D}} & \multicolumn{1}{c|}{50} & \multicolumn{1}{c|}{25} & \multicolumn{1}{c|}{kNN} & \multicolumn{1}{c|}{2} & \multicolumn{1}{c|}{94,67\%} & \multicolumn{1}{c|}{11,08\%} & \multicolumn{1}{c|}{95,39\%} & \multicolumn{1}{c|}{2,57\%} & \multicolumn{1}{c|}{0,73} & \multicolumn{1}{c|}{0,14} \\ \midrule
\multicolumn{1}{|c|}{\textbf{Redondela E}} & \multicolumn{1}{c|}{-} & \multicolumn{1}{c|}{-} & \multicolumn{1}{c|}{kNN} & \multicolumn{1}{c|}{2} & \multicolumn{1}{c|}{97,50\%} & \multicolumn{1}{c|}{7,50\%} & \multicolumn{1}{c|}{94,03\%} & \multicolumn{1}{c|}{4,43\%} & \multicolumn{1}{c|}{0,80} & \multicolumn{1}{c|}{0,16} \\ \midrule
\rowcolor[HTML]{EFEFEF} 
\multicolumn{4}{l}{\cellcolor[HTML]{EFEFEF}\textit{\textbf{Average}}} &  & \textit{\textbf{97,34\%}} & \textit{\textbf{5,41\%}} & \textit{\textbf{91,83\%}} & \textit{\textbf{4,80\%}} & \textit{\textbf{0,75}} & \textit{\textbf{0,14}} \\ \bottomrule
\end{tabular}%
}
\caption{Summary table of the second approach with the models defined as the best in each of the production zones.}
\label{tab:tabla_2}
\end{table}

\section{Discussion} 
\label{sec:d}

The study of the predictor variables for ML models in the prediction of HAB episodes has been one of the most critical points raised in the literature. To date, there is still no consensus on which are the most influential features, varying considerably depending on the geographical region where it is applied and the ML techniques studied. Chlorophyll-a concentration is one of the most relevant features \cite{deng2021machine,yu2021predicting}, as it is directly related to phytoplankton abundance, but in this study, it has been clearly surpassed by the concentration of \textit{D. accuminata}. This is due to the fact that this marker is more accurate when estimating the lipophilic toxin, this dinoflagellate being one of its main producers. It is also necessary to highlight the importance of nutrients such as nitrate and nitrite \cite{yu2021predicting} and environmental factors such as temperature and salinity \cite{yniguez2020predicting}.

The results offered by the kNN machine learning algorithm have been the best for the problem analysed in this work, which is the creation of a predictive model of high toxicity events in mussel production areas (reaching mean sensitivity, mean accuracy and mean kappa index values of 97.34\%, 91.83\% and 0.75 respectively). Its best values of sensitivity, accuracy and kappa have been higher than those obtained with Random forest, ANN, SVM, Na\"ive Bayes and XGBoost techniques (see figure \ref{fig:models}). It should be noted that the average kappa value obtained (0.75) has a substantial degree of agreement according to the scale of values proposed by Landis and Koch (\cite{landis1977measurement}).\\
In the figure \ref{fig:ranking}, it can be seen how the SVM, ANN, Random Forest and XGBoost algorithms are more susceptible than kNN and Na\"ive Bayes to the frequency and duration of mussel harvesting prohibition periods in the production areas. This relationship can be seen in the decrease of the sensitivity values in the areas where these periods are less common (Redondela B, Redondela C and Redondela D), while the values of accuracy remain stable. It is necessary to highlight how the performance of the ANNs tends to offer high values of accuracy and low values of sensitivity for the areas where the state of prohibition of extraction is less common, while in the areas where the number of days of prohibition increases (Cangas F and Cangas G), the model offers an improvement in the values of sensitivity to the detriment of accuracy.\\
These results reflect the imbalance present in the input data which, in areas such as Redondela C, reach a difference of 7\% of positives compared to 93\% of negatives. Therefore, areas such as Cangas F, Cangas G and Cangas H, which have a distribution of closures of around 60-40\%, always obtain better results than areas where FAN is less frequent and where there are fewer cases for the analysis of this study, such as Redondela B, Redondela C and Redondela D, which have a ratio of closures of around 10\%.

\section{Conclusions} 
\label{sec:c}
Although the work carried out to date has obtained good results in predicting biomarkers of FAN, the control of the state of the production areas is conditioned by other external factors, which means that the definition of the problem changes. Some work has used real-time prediction of shellfish and fish mortality events as HAB markers (\cite{yniguez2020predicting}). But real-time prediction does not provide reaction time to these events. However, in this study we have achieved 3-day predictions while maintaining good results. For this we have used the presence of a toxin level above the risk threshold as a HAB marker. In the Galician coast, some previous works seek to solve this problem (\cite{molares2020}), achieving sensitivity and accuracy values of 67.4\% and 83\% in the production area of ``Vigo A'' by applying the ANN technique, while in the present work a significant improvement in the results has been achieved.\\
The approach of the study has shown that it is possible to estimate the status of production areas affected by marine biotoxin events using machine learning techniques. For this purpose, an extensive historical record of variables related to the occurrence of episodes of high toxicity in mussels has been used. The estimates obtained with the models studied have achieved high values of sensitivity and accuracy, so that the expectations initially set out in this study have been met. It has been found that the machine learning algorithm that offers the best results for the resolution of this specific problem in all the production areas of the estuary is the kNN technique. Its best sensitivity and accuracy values have been superior to those obtained with the techniques of Random forest, ANN, SVM, Na\"ive Bayes and XGBoost.\\
The models developed during the study can be used to assess the robustness of the decisions taken by experts when managing the opening or closure of production areas in the absence of recent sampling. This dual assessment mechanism can help experts in complex situations where forecast errors are more likely.

\section{Future Works} 
\label{sec:fw}
In this work, 6 different machine learning algorithms were studied to solve the problem. It is proposed to compare the results obtained with other alternative algorithms that can approach the problem from another perspective, such as hybrid machine learning algorithms (\cite{behera2016ensemble}).\\
The study has focused only on the Vigo estuary, as it is one of the most important Galician estuaries for the production of mussels, and because of its geomorphological characteristics that give it a behaviour in the distribution and evolution of algal blooms of great scientific interest. However, the study is continuing with the aim of supporting the rest of the Galician estuaries with mussel production.\\
In this study, variables identified as relevant in the state of the art have been selected. However, other new variables (e.g. wind, currents, other toxic phytoplankton species, etc.) could be considered as input parameters in the training of machine learning algorithms.\\
One of the limiting factors in conducting this study has been the amount of missing data from the time series used as data sets. It is therefore considered that the creation of a system capable of obtaining or generating (synthetic data (\cite{chen2021synthetic})) such data could lead to a significant improvement in the results obtained.


\section*{Acknowledgments}
The authors want to acknowledge the support from INTECMAR, who have provide mostly data for this work and CESGA, who allows to conduct the tests on their installations. Funding for open access charge: Universidade da Coruña/CISUG. This work is supported by the “Collaborative Project in Genomic Data Integration (CICLOGEN)” PI17/01826 funded by the Carlos III Health Institute in the context of the Spanish National Plan for Scientific and Technical Research and Innovation 2013–2016 and the European Regional Development Funds (FEDER)---"A way to build Europe." This project was also supported by the General Directorate of Culture, Education and University Management of Xunta de Galicia (Ref. ED431G/01, ED431D 2017/16), the “Galician Network for Colorectal Cancer Research” (Ref. ED431D 2017/23), Competitive Reference Groups (Ref. ED431C 2018/49) and the Spanish Ministry of Economy and Competitiveness via funding of the unique installation BIOCAI (UNLC08-1E-002, UNLC13-13-3503) and the European Regional Development Funds (FEDER). Enrique Fernandez-Blanco would also like to thank NVidia corp., which granted a GPU used in this work for the preliminary tests.

\bibliographystyle{unsrt}  
\bibliography{references}  

\begin{appendices}

\begin{table}[h].
  \centering
  \includegraphics[width=0.9\textwidth]{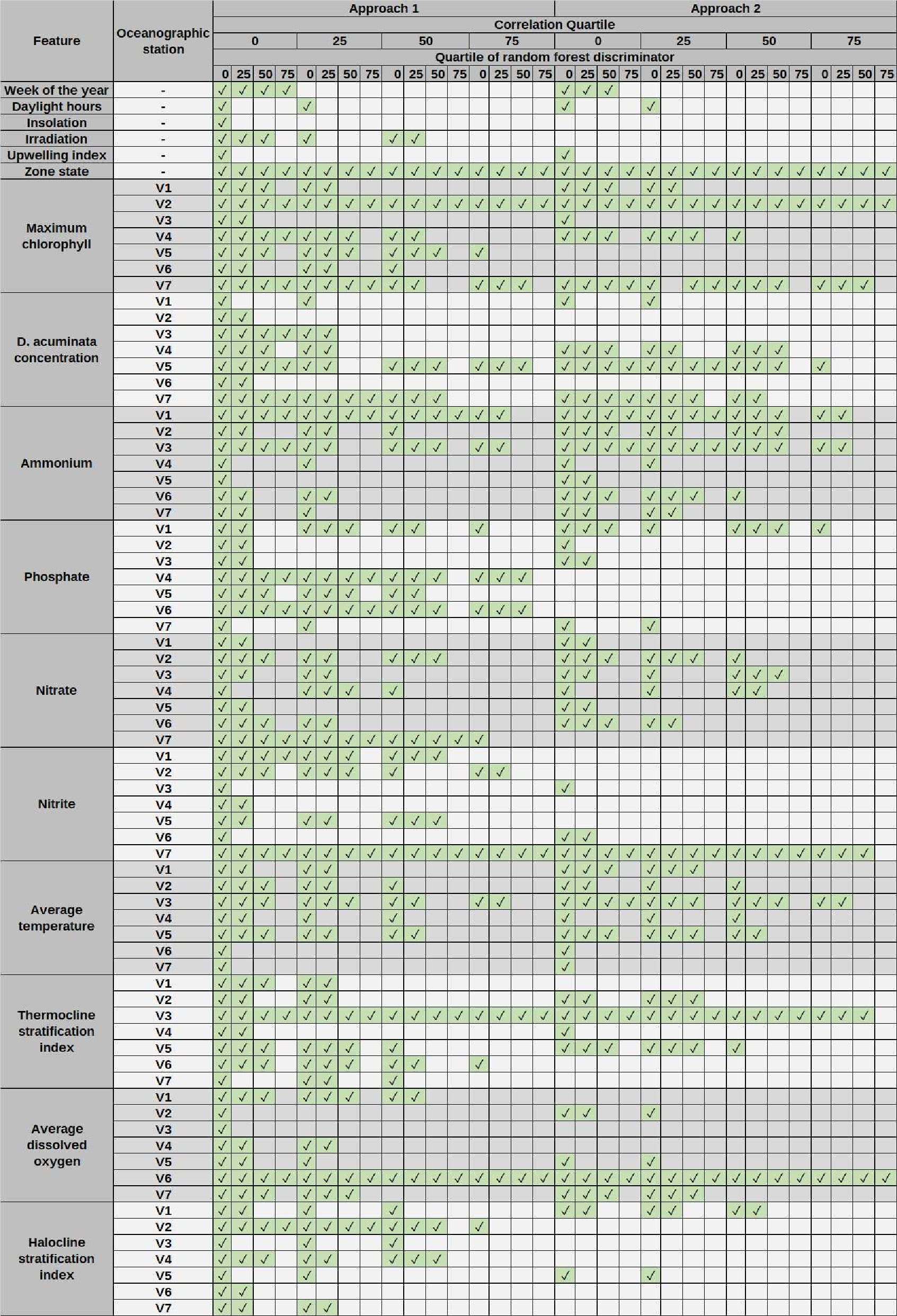}
  \caption{Table with the input features associated with each test block in the Cangas F zone. The check marks when the feature was used}
  \label{fig:rank_cangasF}
\end{table}

\begin{table}[h].
  \centering
  \includegraphics[width=0.9\textwidth]{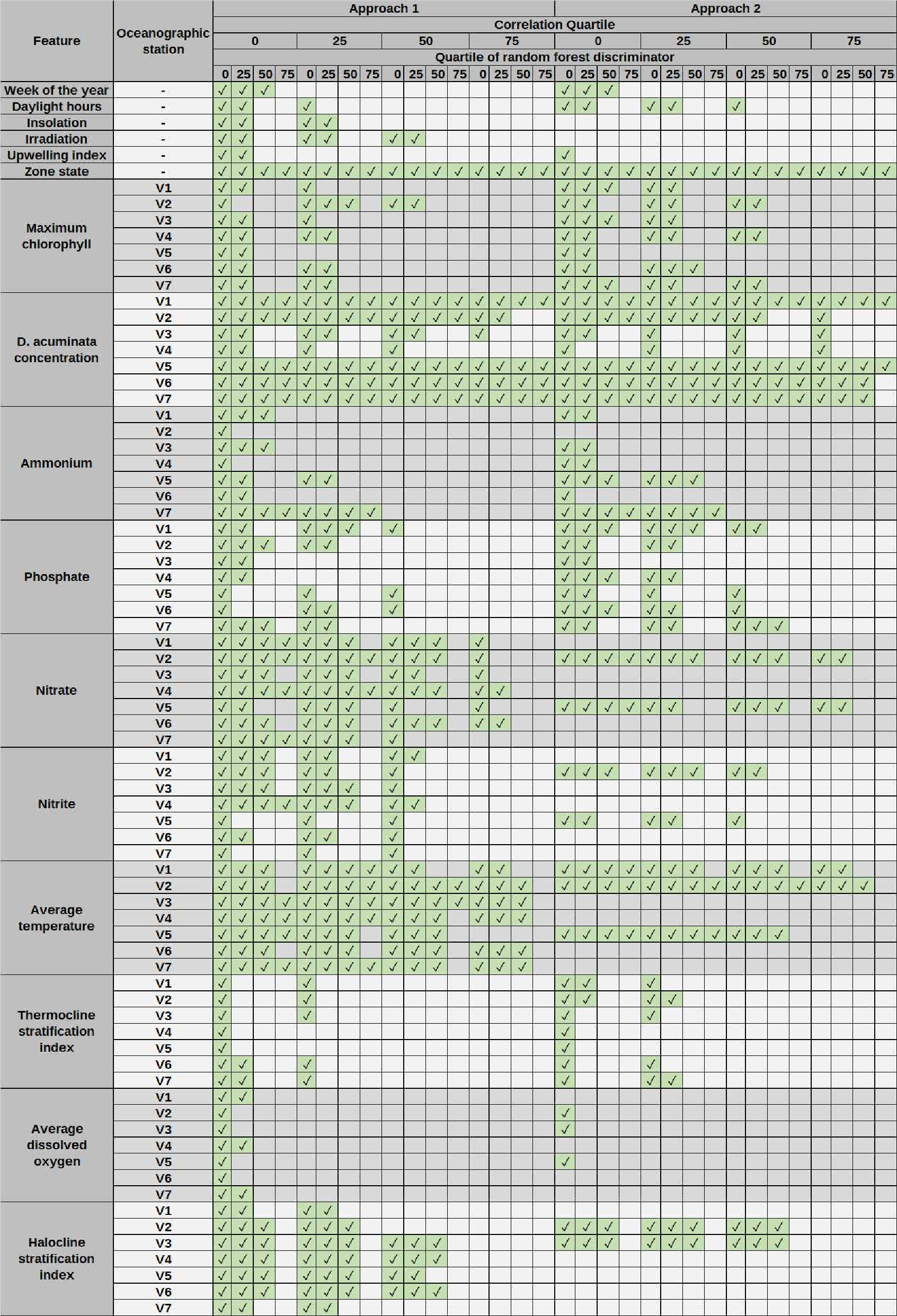}
  \caption{Table with the input features associated with each test block in the Cangas G zone. The check marks when the feature was used}
  \label{fig:rank_cangasG}
\end{table}

\begin{table}[h].
  \centering
  \includegraphics[width=0.9\textwidth]{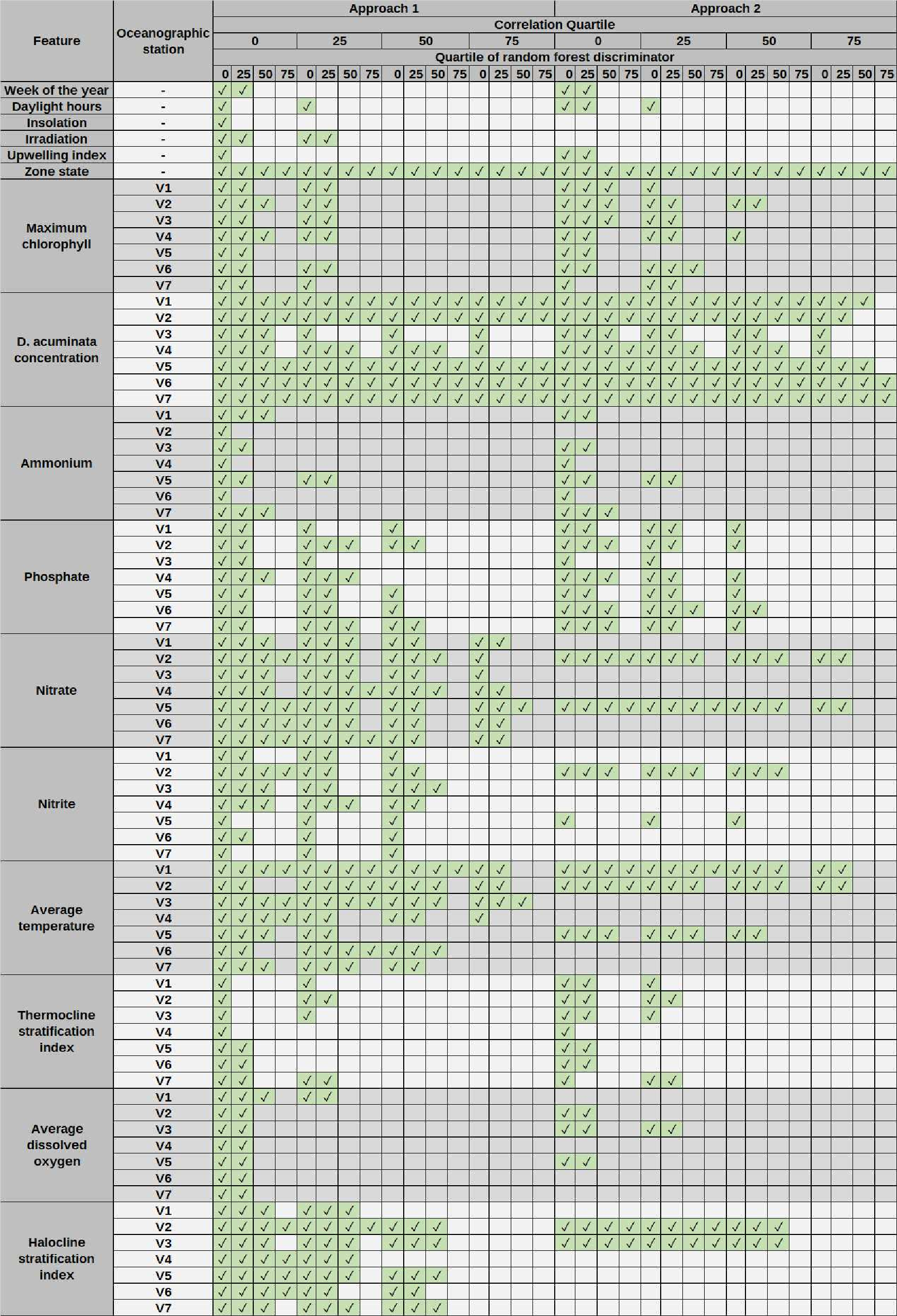}
  \caption{Table with the input features associated with each test block in the Cangas H zone. The check marks when the feature was used}
  \label{fig:rank_cangasH}
\end{table}

\begin{table}[h].
  \centering
  \includegraphics[width=0.9\textwidth]{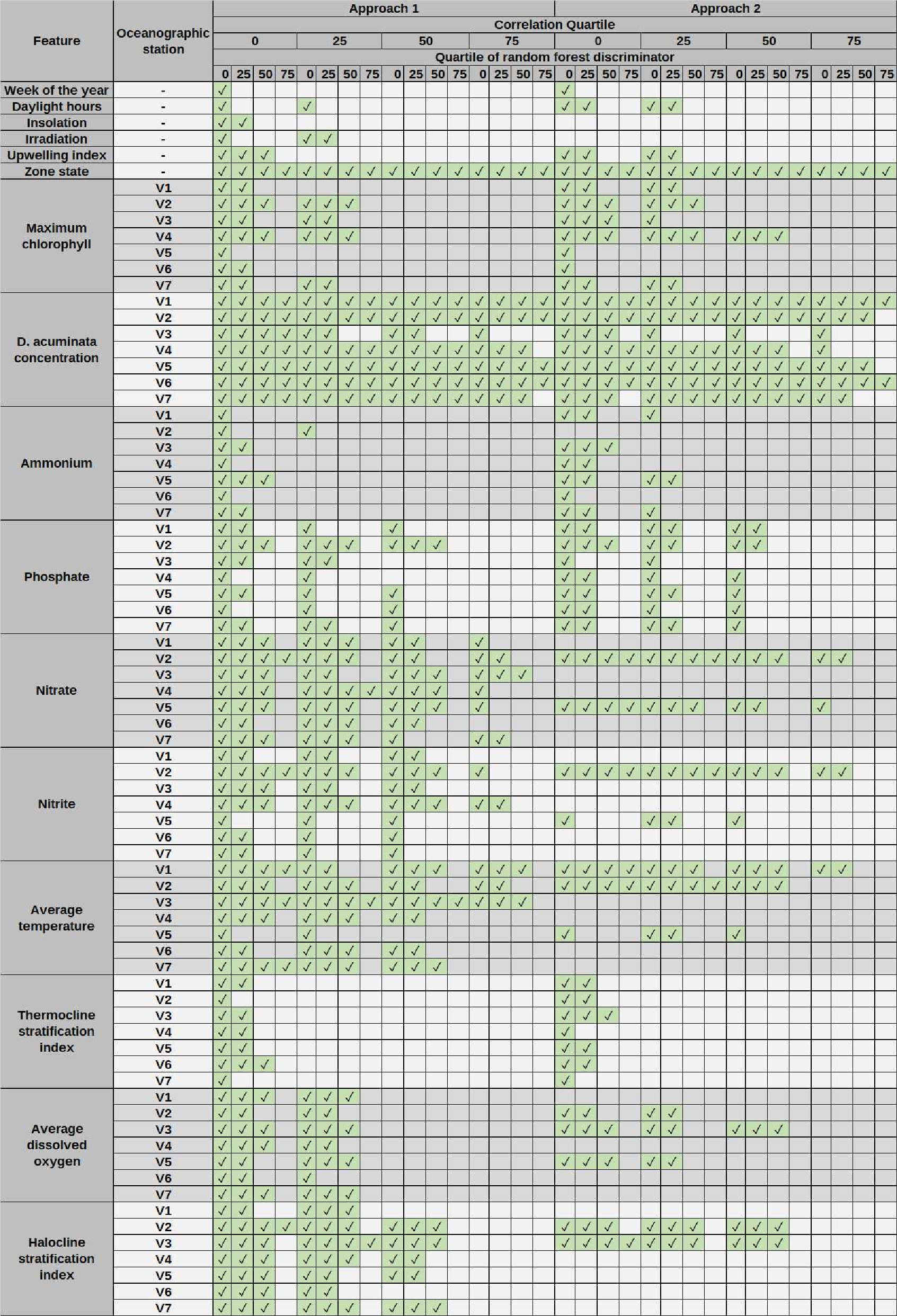}
  \caption{Table with the input features associated with each test block in the Cangas C zone. The check marks when the feature was used}
  \label{fig:rank_cangasC}
\end{table}

\begin{table}[h].
  \centering
  \includegraphics[width=0.9\textwidth]{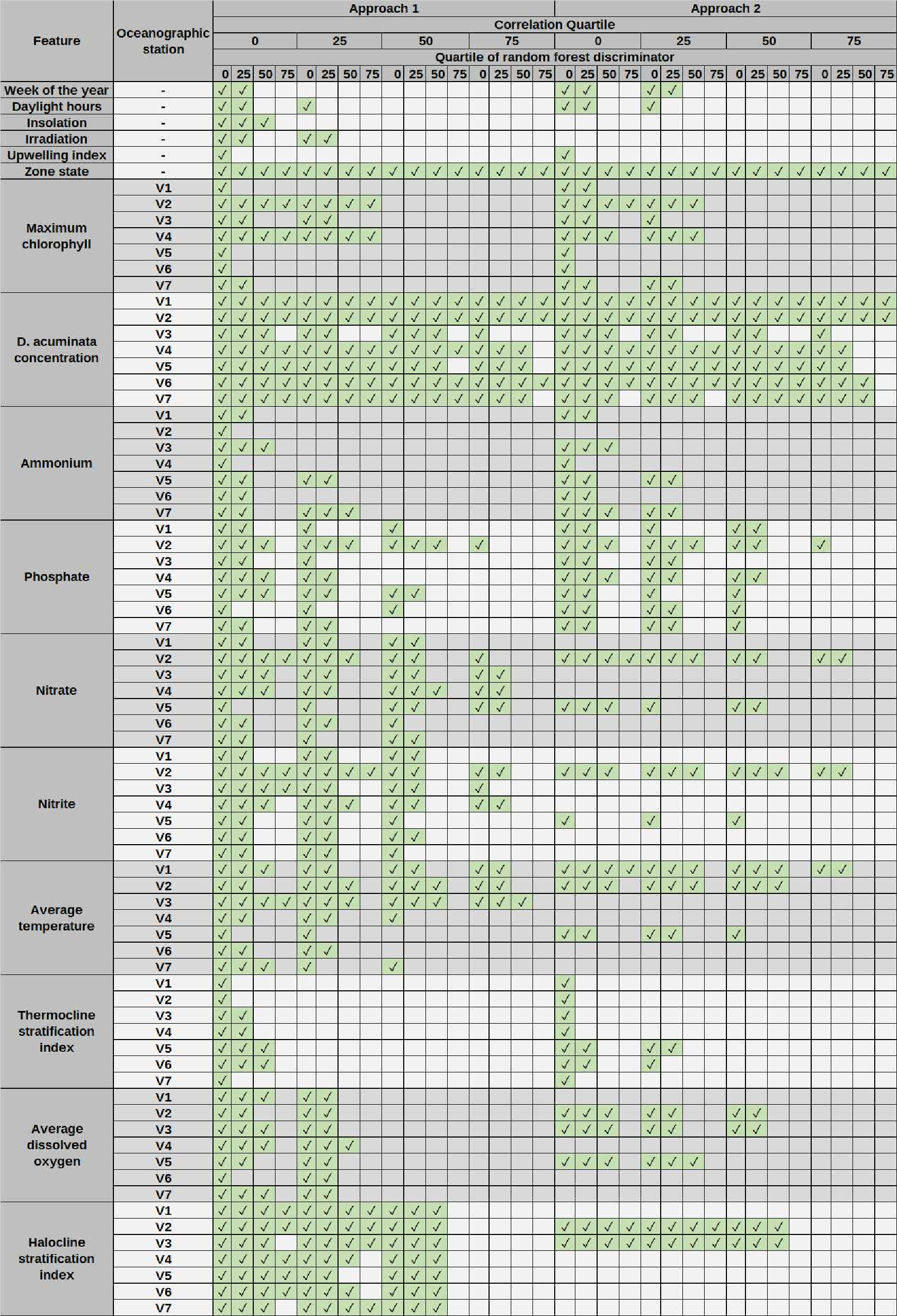}
  \caption{Table with the input features associated with each test block in the Cangas D zone. The check marks when the feature was used}
  \label{fig:rank_cangasD}
\end{table}

\begin{table}[h].
  \centering
  \includegraphics[width=0.9\textwidth]{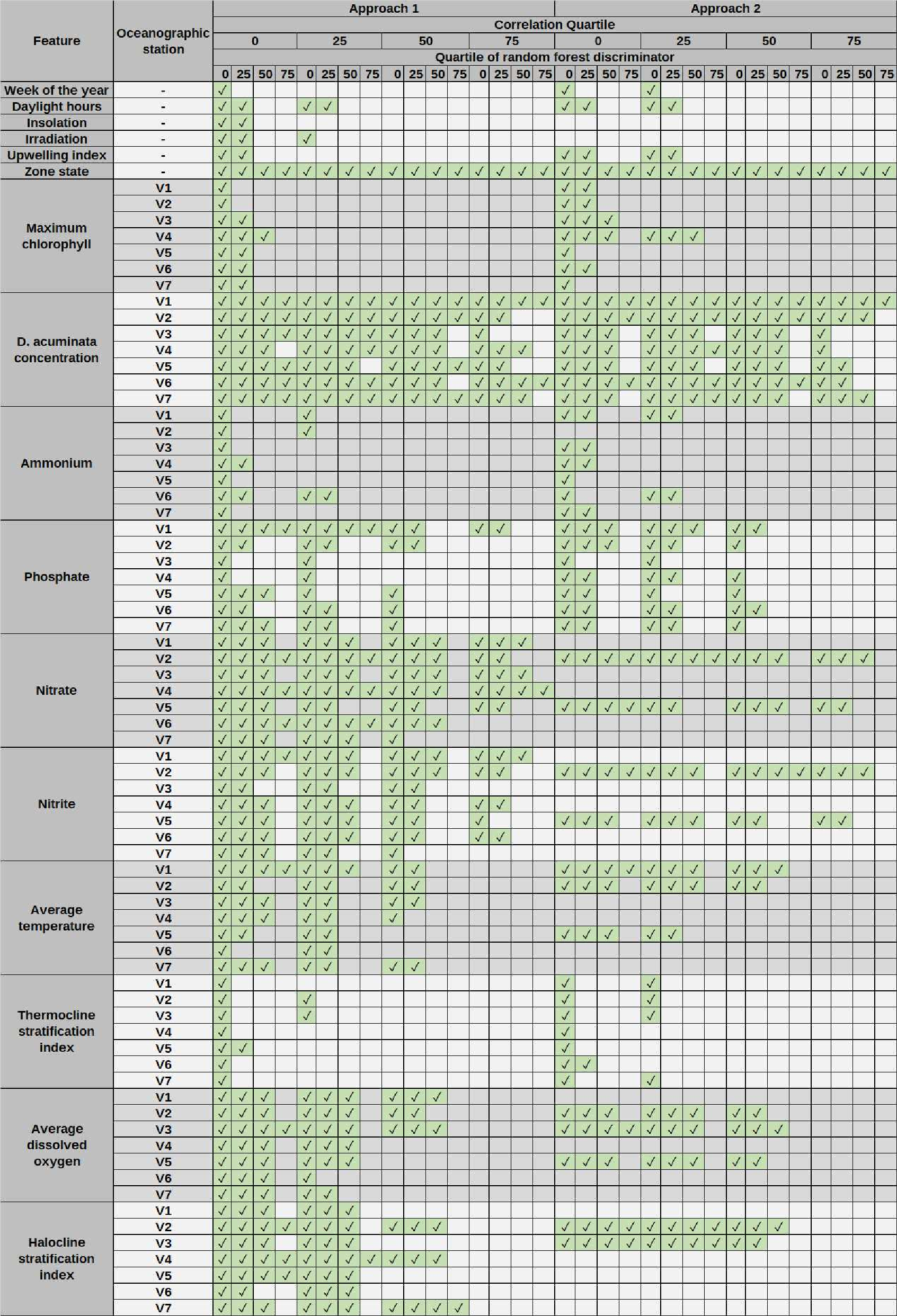}
  \caption{Table with the input features associated with each test block in the Cangas E zone. The check marks when the feature was used}
  \label{fig:rank_cangasE}
\end{table}

\begin{table}[h].
  \centering
  \includegraphics[width=0.9\textwidth]{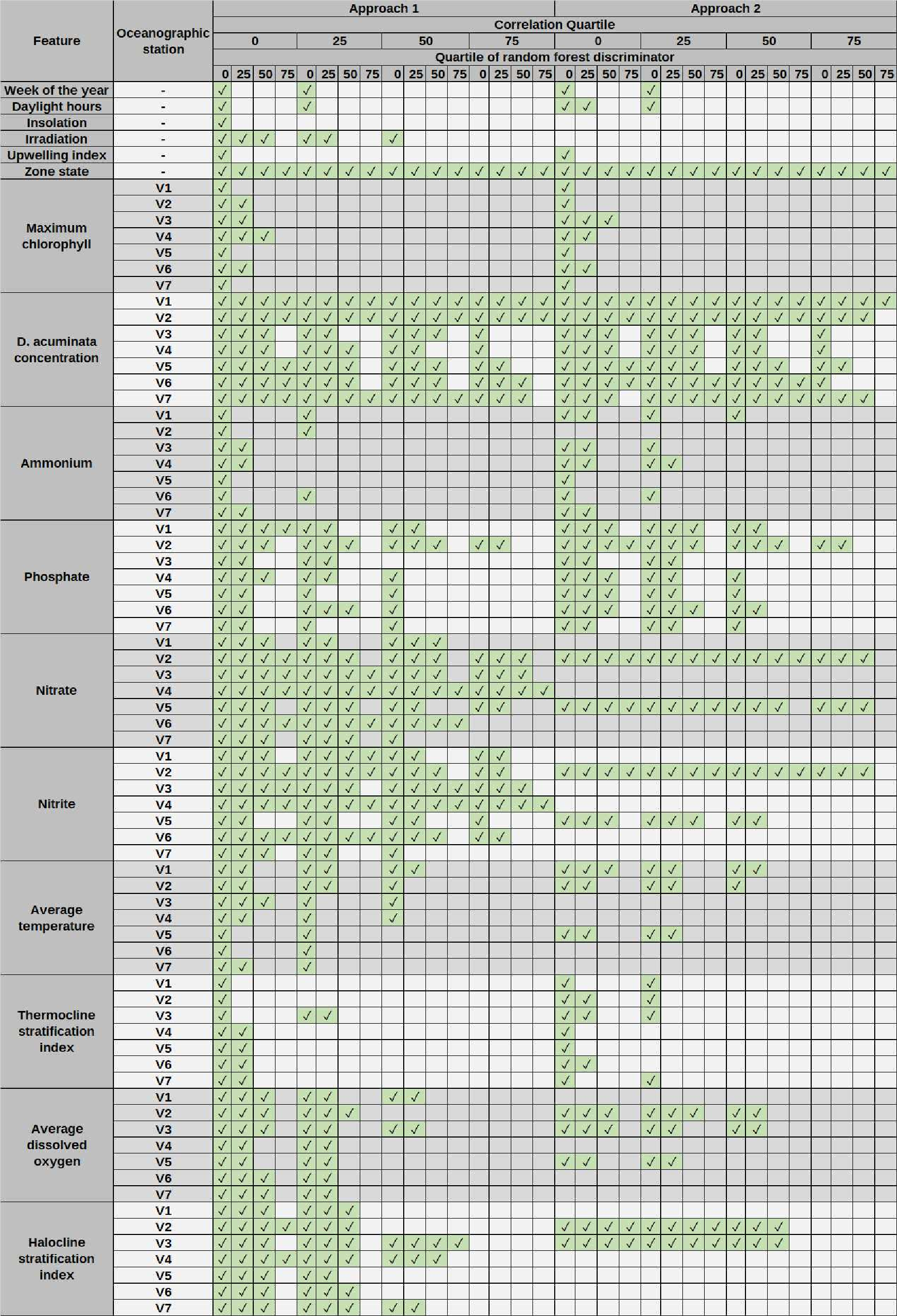}
  \caption{Table with the input features associated with each test block in the Redondela A zone. The check marks when the feature was used}
  \label{fig:rank_redondelaA}
\end{table}

\begin{table}[h].
  \centering
  \includegraphics[width=0.9\textwidth]{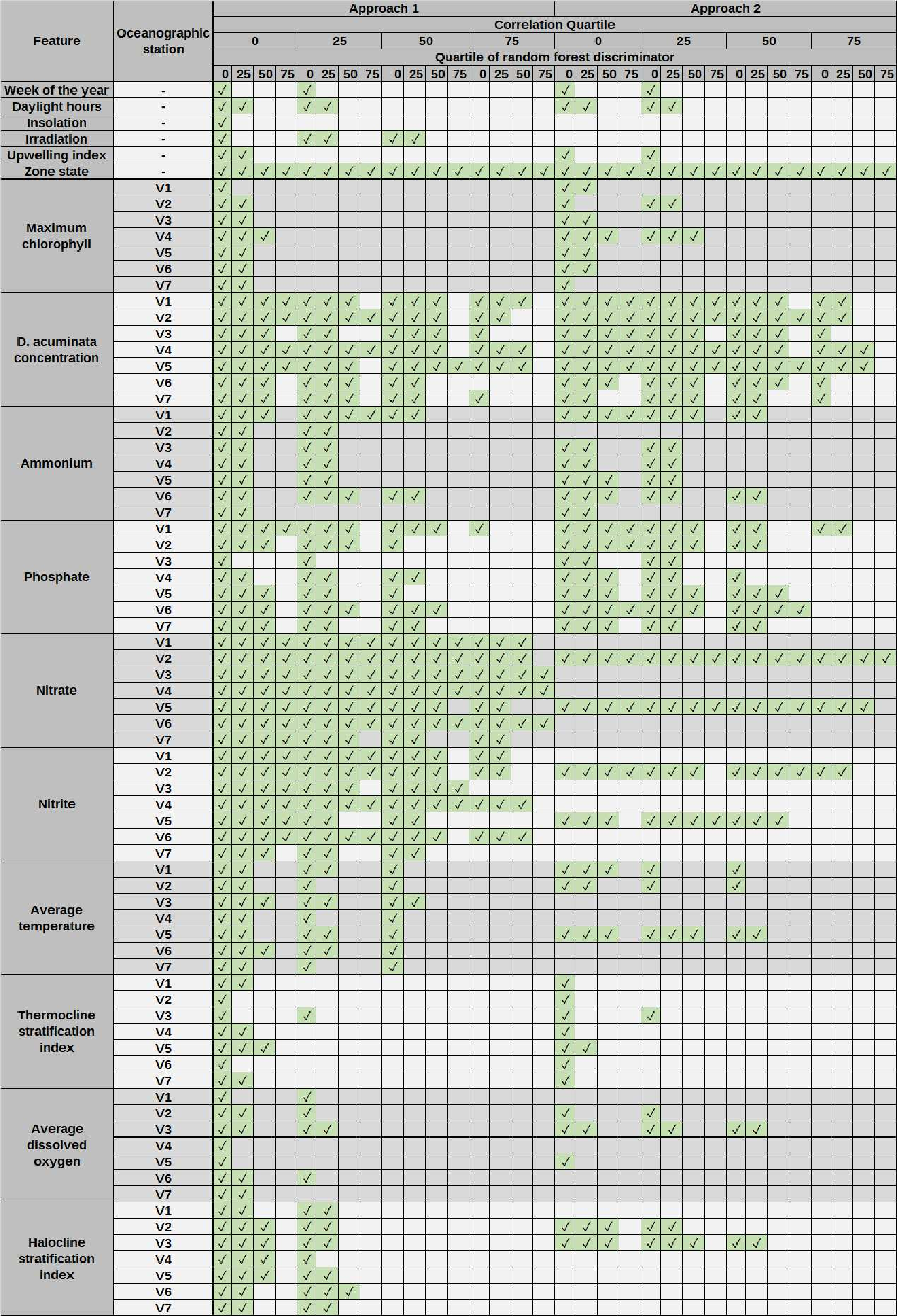}
  \caption{Table with the input features associated with each test block in the Redondela B zone. The check marks when the feature was used}
  \label{fig:rank_redondelaB}
\end{table}

\begin{table}[h].
  \centering
  \includegraphics[width=0.9\textwidth]{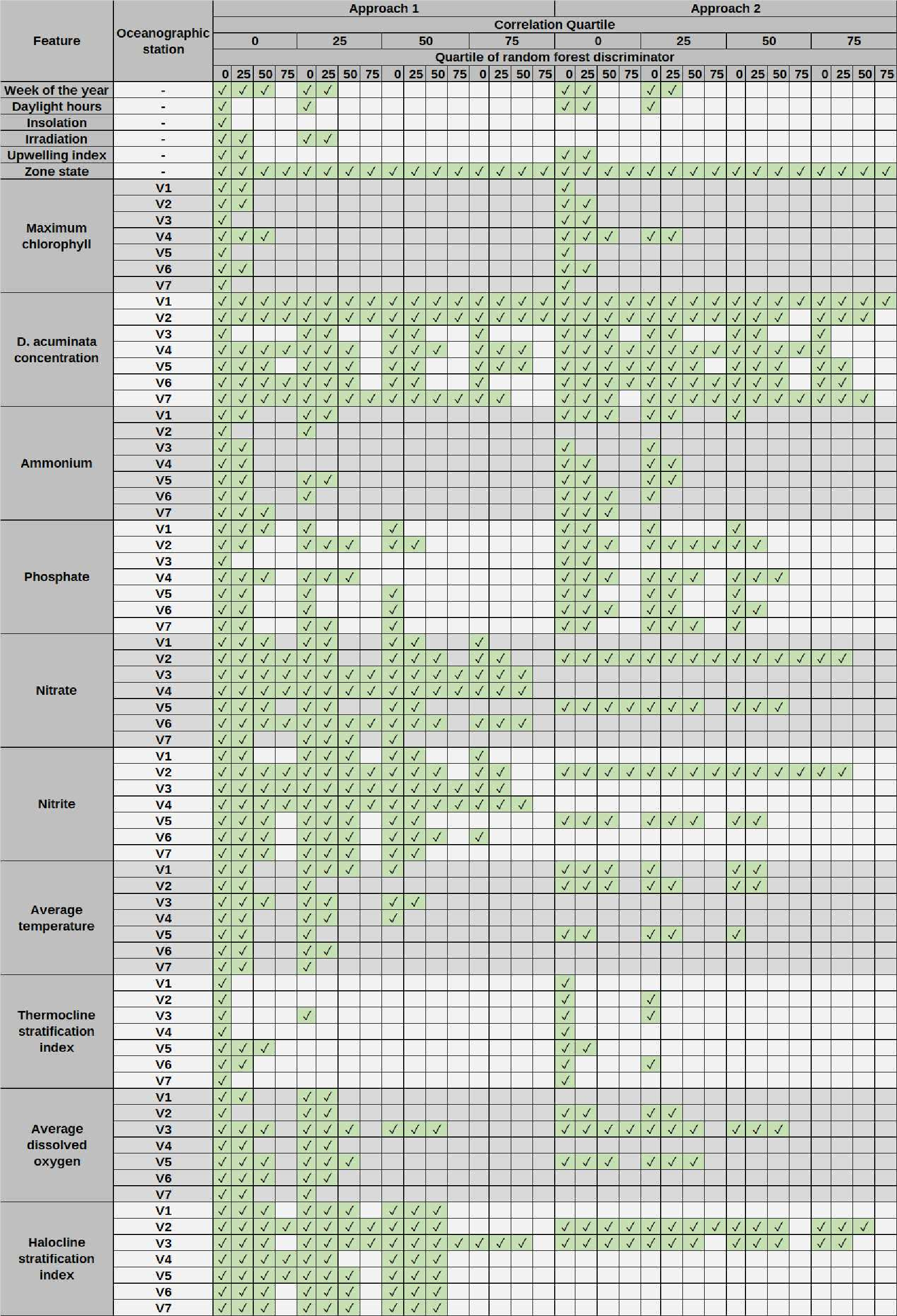}
  \caption{Table with the input features associated with each test block in the Redondela C zone. The check marks when the feature was used}
  \label{fig:rank_redondelaC}
\end{table}

\begin{table}[h].
  \centering
  \includegraphics[width=0.9\textwidth]{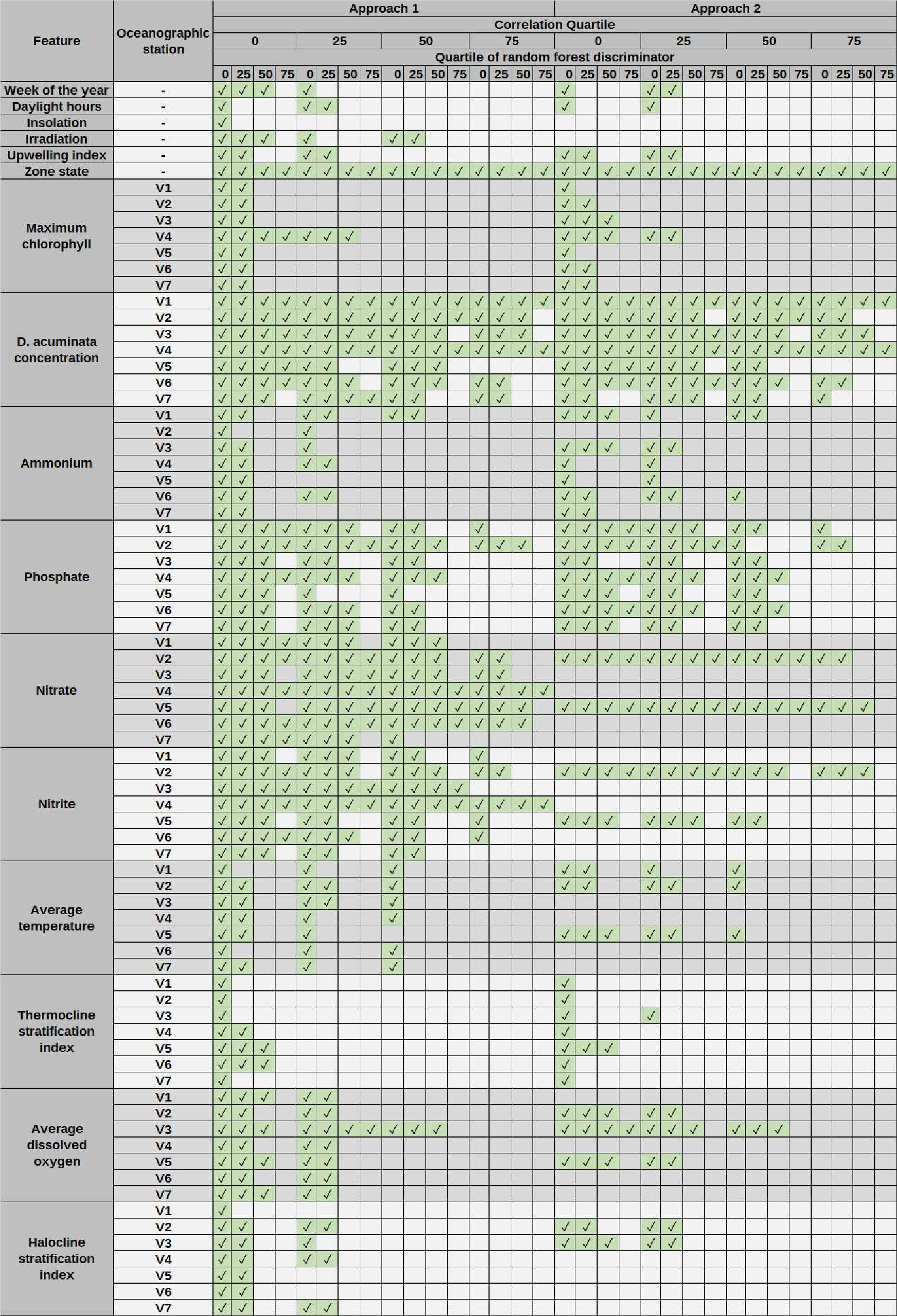}
  \caption{Table with the input features associated with each test block in the Redondela D zone. The check marks when the feature was used}
  \label{fig:rank_redondelaD}
\end{table}

\begin{table}[h].
  \centering
  \includegraphics[width=0.9\textwidth]{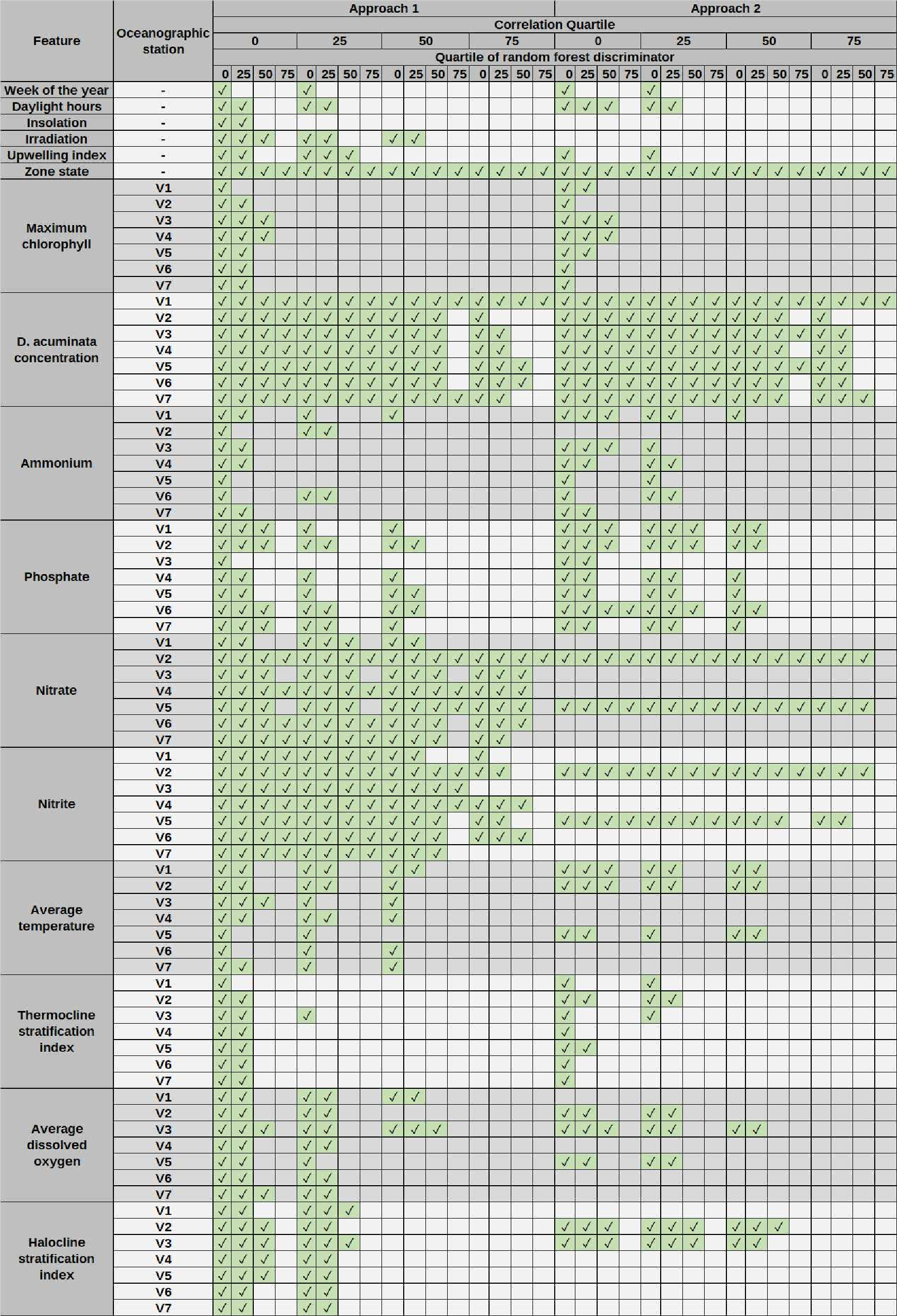}
  \caption{Table with the input features associated with each test block in the Redondela E zone. The check marks when the feature was used}
  \label{fig:rank_redondelaE}
\end{table}

\begin{table}[h].
  \centering
  \includegraphics[width=0.9\textwidth]{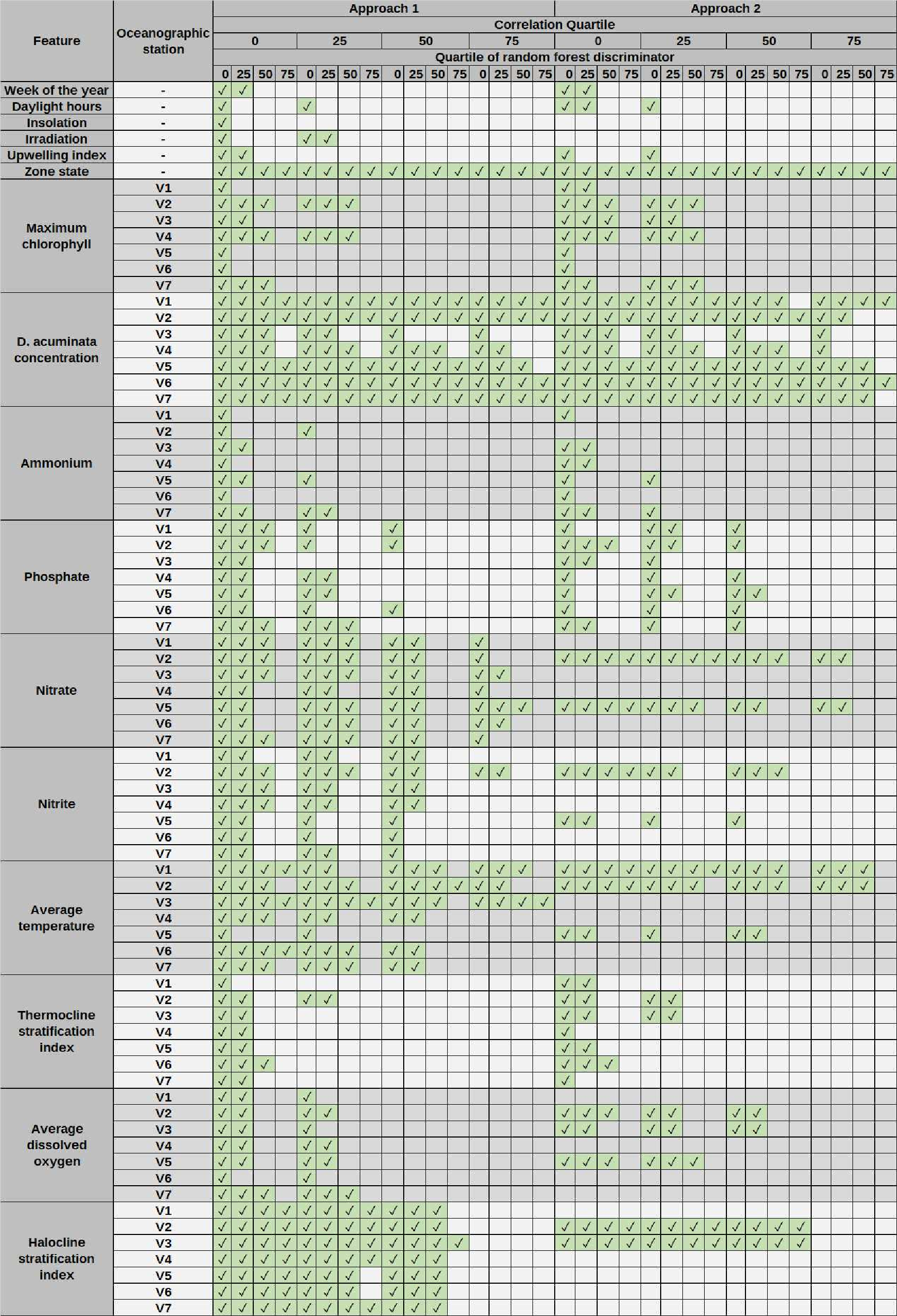}
  \caption{Table with the input features associated with each test block in the Vigo A zone. The check marks when the feature was used}
  \label{fig:rank_vigoA}
\end{table}

\end{appendices}

\end{document}